\documentclass{article}

\usepackage[main, final]{neurips_2026}

\usepackage[utf8]{inputenc}
\usepackage[T1]{fontenc}
\usepackage{url}
\usepackage{booktabs}
\usepackage{amsfonts}
\usepackage{amsmath}
\usepackage{amssymb}
\usepackage{amsthm}
\usepackage{graphicx}
\usepackage{xspace}
\usepackage{microtype}
\usepackage{multirow}
\usepackage{wrapfig}
\usepackage{algorithm}
\usepackage{algpseudocode}
\usepackage[dvipsnames]{xcolor}
\usepackage{xcolor}     
\usepackage{longtable}  
\usepackage{pdflscape}  

\usepackage{hyperref}       

\usepackage[normalem]{ulem}
\usepackage{xcolor}

\newcommand{\method}{\textsc{StepBack}\xspace}
\newcommand{\ourmodel}{Qwen2.5-3B-Instruct\xspace}
\newcommand{\gsm}{GSM8K\xspace}
\newcommand{\mathfive}{MATH500\xspace}

\title{\textbf{StepBack}: Step-Level Error-Localized Resampling for Efficient Test-Time Reasoning}

\author{
  Yang Ouyang \qquad \qquad Jung-Eun Kim\thanks{Corresponding Author} \\
  Department of Computer Science\\
  North Carolina State University \\
  \texttt{\{youyang7, jung-eun.kim\}@ncsu.edu}
}

\begin{document}

\maketitle

\begin{abstract}
Self-consistency (SC) is a widely used decoding strategy for improving chain-of-thought reasoning, but it requires sampling many complete reasoning trajectories with a preset budget. Its efficient variants reduce the cost mainly by stopping full-trajectory sampling early or distributing the number of samples across questions. In this paper, we propose \textbf{StepBack}, a step-level error-localized resampling method for efficient test-time reasoning. Given a small set of draft trajectories, \method localizes likely first error positions using lightweight signals and then resamples the detected error step and its suffix, followed by majority answer voting. Empirically, \method matches the accuracy of SC on average across seven mathematical reasoning and multi-hop QA benchmarks, while reducing token usage by up to 41.26\%. In addition, by nature, \method can be combined with adaptive inference methods such as Adaptive Consistency, further improving their inference-time efficiency. \footnote{Our code is available at: \url{https://github.com/JEKimLab/StepBack}}

\end{abstract}

\section{Introduction}
\begin{wrapfigure}[20]{r}{0.45\textwidth}
    \centering
    \vspace{-0.1in}
    \includegraphics[width=\linewidth]{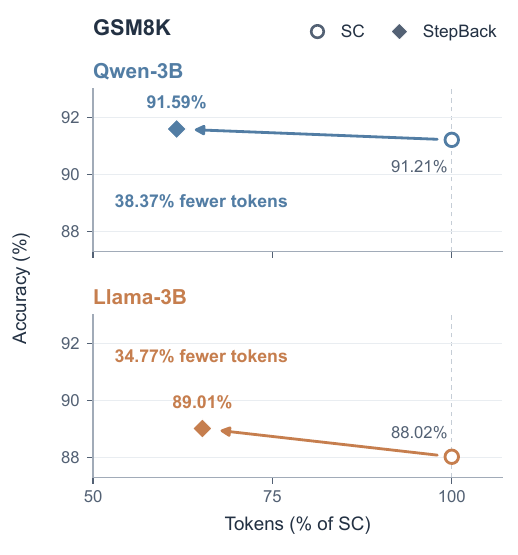}
    \caption{\method achieves higher accuracy with significantly fewer generated tokens than self-consistency on GSM8K.}
    \vspace{-0.1in}
    \label{fig:motivation}
\end{wrapfigure}

Chain-of-thought prompting improves the reasoning ability of large language models (LLMs) \citep{brown2020fewshot}  by encouraging language models to verbalize intermediate steps before producing a final answer \citep{wei2022chain,kojima2022large}. A widely used approach to further improve accuracy is Self-Consistency (SC), which samples multiple ``complete'' reasoning trajectories and selects the most frequent final answer \citep{wang2022self}. Despite its effectiveness, SC is computationally expensive: each additional candidate answer requires generating an entire reasoning trajectory from scratch, so the token cost grows linearly with both the number of sampled trajectories and the length of each trajectory.

\begin{figure}
    \centering
    \includegraphics[
    width=\linewidth,
    trim={5mm 0 0 0},
    clip
]{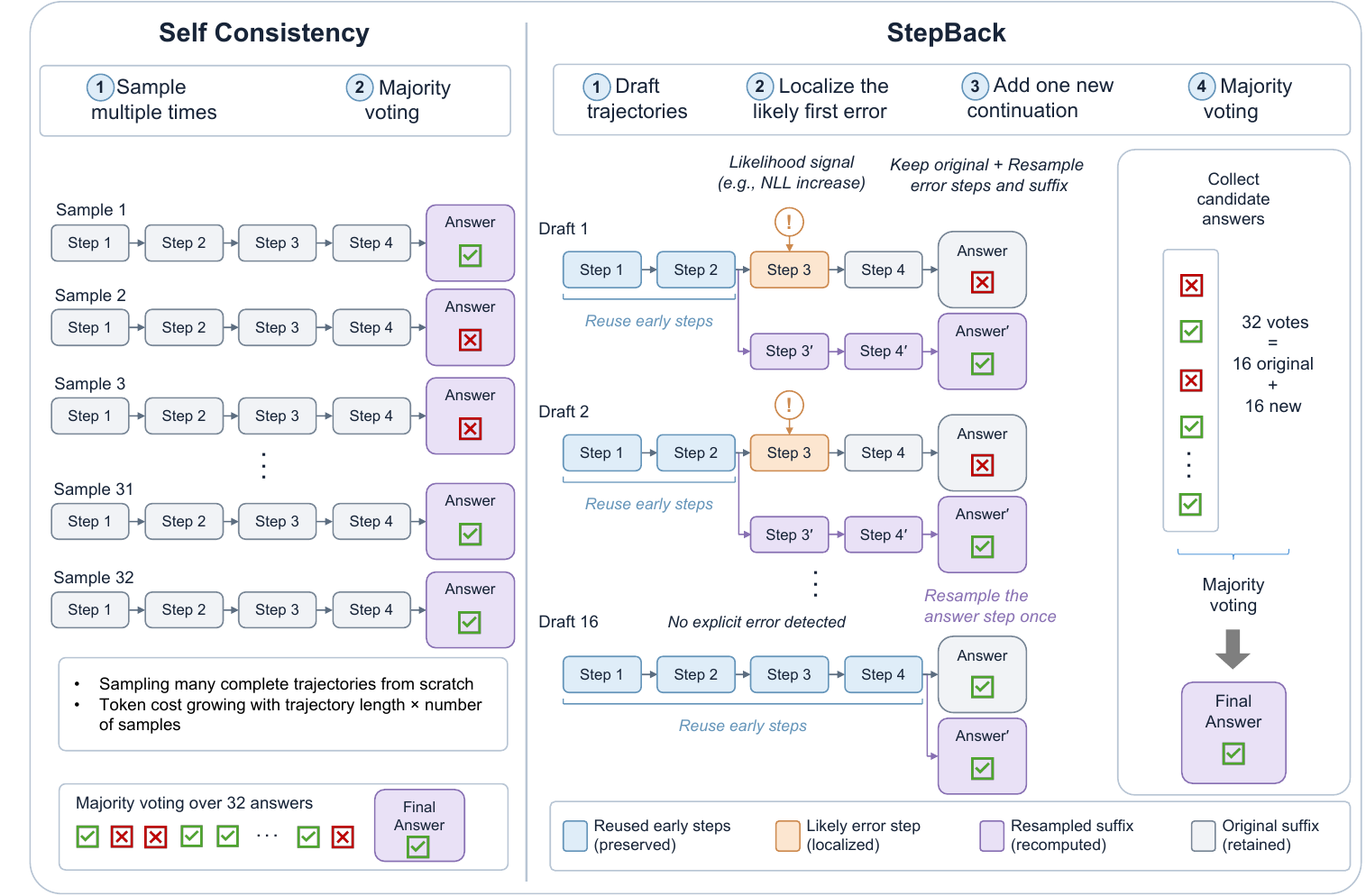}
    \caption{Instead of regenerating many full trajectories, StepBack identifies a likely first error step, reuses valid early steps, and resamples only the erroneous step and its suffix before majority voting with the same voting budget (32 in this example).}
    \label{fig:overview}
\end{figure}

Recent studies have sought to reduce this cost by adjusting the number of full trajectories to sample. Adaptive-Consistency (AC) and Early-Stopping Self-Consistency (ESC) stop sampling once sufficient answer agreement is observed; Difficulty-Adaptive Self-Consistency (DSC) allocates larger budgets to harder questions; Reasoning-Aware Self-Consistency (RASC) makes stopping and voting more selective with rationale features~\citep{aggarwal2023lets,li2024escape,wang2025penny,wan2025reasoning}. These methods improve decisions on \emph{when} to stop sampling or \emph{how many} trajectories to sample, but they largely operate at the level of complete trajectories. As a result, they leave open a finer-grained question: \emph{where to efficiently spend additional sampling within a reasoning trajectory?}

This question is important because reasoning failures often emerge at an intermediate reasoning step rather than at only the final answer \citep{wei2022chain, uesato2022solving, lightman2023lets}. Once such an error occurs, subsequent steps may hinge on the corrupted reasoning state, causing the error to propagate and end up with an incorrect final answer, a phenomenon also noted in prior work on accumulated errors in chain-of-thought reasoning \citep{ling2023deductive, ouyang2026densesteer}. Meanwhile, the early steps of a reasoning chain contain useful problem setup \citep{ji2025first}. This suggests a sensible opportunity for efficient test-time reasoning: if a likely first erroneous step can be localized, the valid early steps can be preserved while only the erroneous step and its subsequent suffixes are resampled \citep{tyen2024llms}. 

With this motivation, we propose \method, a step-level error-localized resampling method for efficient self-consistent reasoning. \method first generates a small set of draft trajectories and decomposes each trajectory into reasoning steps. It then uses lightweight signals, such as step-level negative log-likelihood, to localize likely first error positions. For each localized trajectory, \method reuses the earlier steps before the detected error, rolls back to the error step, and resamples the detected error step together with its suffixes. When no explicit error is detected, \method samples only the final answer step. The resulting candidate answers are then aggregated by majority voting, following the standard SC. This approach significantly cuts token usage while matching SC's accuracy on average, as depicted in Figure~\ref{fig:motivation}. The key contributions of our work are:

\newpage
\paragraph{Contributions.}
\begin{itemize}
    \item We introduce \method, an efficient step-level error-localized resampling method that reuses draft early steps, localizes likely first error positions with lightweight signals, and performs suffix resampling instead of repeated full-trajectory generation.

    \item To the best of our knowledge, \method is the first self-consistency-based decoding method that uses step-level error localization to decide where to begin resampling, rather than sampling every candidate as a full trajectory.

    \item We empirically show across mathematical reasoning and multi-hop QA benchmarks that \method can match or surpass full self-consistency while using substantially fewer generation tokens. We further show that \method can be seamlessly combined with adaptive inference methods such as Adaptive Consistency, Early-Stopping SC, and Difficulty-Adaptive SC.
\end{itemize}

\section{Related Work}

\paragraph{Efficient self-consistency.}
Recent work reduces self-consistency cost by adaptively controlling the number of full trajectories to sample. Adaptive-Consistency stops once answer agreement is sufficient \citep{aggarwal2023lets}; Early-Stopping Self-Consistency uses simple window-based stopping rules \citep{li2024escape}; Reasoning-Aware Self-Consistency incorporates rationale-quality features into stopping and voting \citep{wan2025reasoning}; and Difficulty-Adaptive Self-Consistency allocates larger budgets to harder questions \citep{wang2025penny}. These methods improve sampling efficiency by deciding \emph{how many} trajectories to generate. In contrast, \method solves the problem \emph{where} additional computation should be spent inside a trajectory. Thus, \method is complementary to adaptive SC methods: they reduce the number of consumed candidates, while \method reduces the cost of each candidate.

\paragraph{Step-level signals and backtracking.}
Process supervision and step-level verification show that intermediate reasoning steps carry useful signals about solution quality \citep{uesato2022solving,lightman2023lets}. Errors tend to propagate once an early step becomes invalid \citep{ling2023deductive}, early prefixes often contain reusable problem setup \citep{ji2025first}, and LLMs correct reasoning more effectively when the error location is given \citep{tyen2024llms}. Several methods act on such signals by returning to an earlier step and regenerating from there: LeCo~\citep{yao2024correctness} restarts after the lowest-confidence step until two consecutive answers agree; \citet{singh2025improving} backtrack with a step-level verifier conditioned on previously attempted revisions; Adaptive Rectification Sampling~\citep{tan2025adaptive} prompts the model to rethink a step whose PRM score falls below a threshold; Thought Rollback~\citep{chen2024toward} rolls back to earlier thoughts after analyzing their errors; and Self-Backtracking~\citep{yang2026step} trains the model itself to decide when and where to backtrack. In these methods, each rollback depends on the outcome of earlier attempts, so revisions run sequentially and their total cost is not fixed in advance. \method instead uses lightweight signals, such as a negative log-likelihood increase or a process-reward score drop, to assign a single rollback point to each independent draft, so all suffix continuations can be generated in parallel and the number of candidate answers is fixed at \(B=n_d n_s\), the same budget as full-trajectory self-consistency. Each additional candidate reuses its draft prefix and therefore costs fewer tokens than a full trajectory, and \method requires no training and aggregates answers by standard majority voting.
\section{Preliminaries and Motivating Observations}
\label{sec:prelim}

\subsection{Step-Level Reasoning Trajectories}
\label{sec:step_view}

Let \(x\) be an input problem and \(p_\theta\) be the language model. We write a reasoning trajectory as
\begin{equation}
    \tau = (z_1, z_2, \ldots, z_T, z_{T+1}),
\end{equation}
where \(z_1,\ldots,z_T\) are intermediate reasoning steps and \(z_{T+1}\) is the final-answer step. We denote the extracted answer by \(a(\tau)\).

Our method relies on the fact that instruction-tuned models usually produce naturally separable reasoning steps. For example, Qwen2.5-3B-Instruct often separates steps with blank lines, i.e., ``\texttt{\textbackslash n\textbackslash n}'', in both math and multi-hop QA tasks. The other models we test show similar line-delimited outputs.

We therefore parse each trajectory into steps using blank-line boundaries. Examples of model outputs and the resulting step decomposition are provided in Appendix~\ref{app:step_examples}.

\subsection{The Value of Accurate Error Localization}
\label{sec:oracle_error_localization}

For an incorrect trajectory, let \(e(\tau)\) denote the position of the first step where the reasoning becomes invalid.\footnote{When there is no ambiguity, we omit $(\tau)$.} The earlier reasoning steps \(z_{<e(\tau)}\) remain valid, while the subsequent steps \(z_{\ge e(\tau)}\) may be corrupted by the first error.

\paragraph{Draft--suffix resampling.}
Given an answer budget \(B\), we consider a two-stage procedure shared by the oracle study below and by our method (Section~\ref{sec:method}). We first sample \(n_d\) draft trajectories \(\tau^{(1)},\ldots,\tau^{(n_d)}\). For each draft \(\tau^{(j)}\) with \(T_j\) reasoning steps, we choose a rollback point \(e_j\in\{1,\ldots,T_j+1\}\), keep the prefix \(z^{(j)}_{<e_j}\), and draw \(n_s-1\) suffixes
\begin{equation}
    \tilde z^{(j)}_{e_j:T_j+1} \sim p_\theta(\cdot \mid x, z^{(j)}_{<e_j}).
\end{equation}
Together with the draft's own answer, this yields \(B=n_d n_s\) candidate answers, aggregated by majority vote as in self-consistency (SC). The only design choice is how to select \(e_j\).

\paragraph{Oracle rollback.}

To understand the value of accurate error localization, we first consider an oracle setting where the true first-error position, \(e(\tau)\), is known, and set \(e_j=e(\tau^{(j)})\) for each draft. We obtain these oracle positions from first-error annotations produced by an LLM verifier prompt. We manually inspect 100 annotated samples and find no obvious misjudgments; the full annotation protocol and examples are provided in Appendix~\ref{app:first_error_annotation} and Appendix~\ref{app:first_error_example}. 
For drafts without an annotated first error, we set \(e_j=T_j+1\) and resample only the final-step answer.

Table~\ref{tab:oracle_error_localization_main} reports motivating results under this oracle setting. Oracle first-error localization enables suffix resampling to match or even outperform full self-consistency while using fewer tokens. This suggests that many failed trajectories still contain useful early reasoning steps. Therefore, by reliably localizing the first errors and simply resampling from them, we can achieve both token efficiency and comparable (or better at times) accuracy.

\begin{table}[t]
\centering
\small
\caption{
Oracle first-error localization: suffix resampling starts from annotated first-error positions, with \(n_d\) drafts and \(n_s\) answers per draft (\(n_d n_s=32\)). Savings are computed relative to SC@32.
}
\label{tab:oracle_error_localization_main}
\resizebox{0.95\columnwidth}{!}{%
\begin{tabular}{llccc}
\toprule
Dataset & Method & Accuracy & Tokens / Q & Savings \\
\midrule
GSM8K
& SC@32 & 91.21\% & 9506 & --  \\
& Oracle error-localized resampling \(n_d=8,n_s=4\) & 91.89\% & 3427 & 64.0\% \\
& Oracle error-localized resampling \(n_d=16,n_s=2\) & 92.27\% & 5449 & 42.7\% \\
\midrule
MATH500
& SC@32 & 72.80\% & 18335 & -- \\
& Oracle error-localized resampling \(n_d=8,n_s=4\) & 72.40\% & 9815 & 46.5\% \\
& Oracle error-localized resampling \(n_d=16,n_s=2\) & 73.00\% & 12615 & 31.2\% \\
\bottomrule
\end{tabular}%
}
\end{table}

\section{\method: Step-Level Error-Localized Resampling}
\label{sec:method}

The oracle study in Section~\ref{sec:oracle_error_localization} shows that localized resampling can be effective when the first erroneous reasoning step is known. In practice, however, the true first-error position $e(\tau)$ is unavailable at inference time. \method turns this oracle diagnostic into a practical decoding procedure by replacing $e(\tau)$ with a predicted rollback point obtained from lightweight step-level signals.

For each draft $\tau^{(j)}=(z^{(j)}_1,\ldots,z^{(j)}_{T_j+1})$, $j=1,\ldots,n_d$, \method applies an inference-time localization rule,
\begin{equation}
    \hat{e}_j = \mathrm{Loc}(x,\tau^{(j)}) \in \{1,\ldots,T_j+1\}.
\end{equation}
Here $\hat{e}_j \le T_j$ means that step $\hat{e}_j$ is predicted as the first unreliable reasoning step. \method then preserves the early steps $(z^{(j)}_1,\ldots,z^{(j)}_{\hat{e}_j-1})$ and resamples the detected step together with its suffix. If $\hat{e}_j=T_j+1$, no explicit reasoning error is detected, so \method preserves the full reasoning chain and resamples only the final answer step to simply obtain the same number of answers as SC does. The resulting $B$ answer candidates are aggregated with the same normalized majority vote as done in SC. For details, see Algorithm~\ref{alg:method}.

\subsection{Inference-Time Localization Signals}
\label{sec:localization_signals}
We instantiate \(\mathrm{Loc}\) with lightweight inference-time signals in our experiments, including \textsc{NLL-signal} and \textsc{PRM-signal}. Throughout this subsection, we consider a single draft and omit the superscript $(j)$ (unless any confusion arises), writing $\tau=(z_1,\ldots,z_{T+1})$.

\paragraph{NLL-signal: Step-Level Negative Log-Likelihood (NLL) Increase}

For each reasoning step $z_t$, we compute its mean token negative log-likelihood
under the base model:
\begin{equation}
    u_t
    =
    -\frac{1}{|z_t|}
    \sum_{y_i\in z_t}
    \log
    p_\theta
    \left(
    y_i
    \mid
    x,z_{<t},y_{<i}
    \right).
\end{equation}
A sudden increase in $u_t$ indicates that the model becomes less confident when generating the current step. We define
\begin{equation}
    \Delta^{\mathrm{NLL}}_t
    =
    u_t-u_{t-1},
\end{equation}
and predict the rollback point as
\begin{equation}
    \hat{e}_{\mathrm{NLL}}
    =
    \begin{cases}
    \min\{t:\Delta^{\mathrm{NLL}}_t\ge \delta_{\mathrm{NLL}}\},
    & \text{if such a step exists},\\
    T+1,
    & \text{otherwise}.
    \end{cases}
\end{equation}
The fallback position $T+1$ means that no explicit reasoning error is detected. In this case, \method keeps the full reasoning chain and resamples only the final answer step.

\paragraph{PRM-signal: Step-Level Process Reward Model (PRM) Score Drop}

We also consider an external process reward model (PRM), denoted by $R_\phi$. Given the problem $x$ and the partial reasoning trace $z_{\le t}$, the PRM assigns a score
\begin{equation}
    r_t
    =
    R_\phi(x,z_{\le t}),
\end{equation}
which estimates the reliability of the reasoning prefix. A large drop after
adding step $z_t$ suggests that the new step makes the reasoning less reliable.
We define
\begin{equation}
    \Delta^{\mathrm{PRM}}_t
    =
    r_{t-1}-r_t,
\end{equation}
and predict the rollback point as
\begin{equation}
    \hat{e}_{\mathrm{PRM}}
    =
    \begin{cases}
    \min\{t:\Delta^{\mathrm{PRM}}_t\ge \delta_{\mathrm{PRM}}\},
    & \text{if such a step exists},\\
    T+1,
    & \text{otherwise}.
    \end{cases}
\end{equation}
In our experiments, unless otherwise specified, we use Qwen2.5-Math-PRM-7B~\citep{prmlessons} as the PRM scorer.

Both signals are inexpensive. \textsc{NLL-signal} uses token probabilities already available from the base model, while the \textsc{PRM-signal} only scores partial reasoning traces rather than requiring additional full-trajectory sampling. We provide the detailed diagnostics in Appendix~\ref{app:localization_details}.

\subsection{Comparison with Self-Consistency}
\label{sec:comparison_sc}

Standard self-consistency (SC) samples $B$ complete reasoning trajectories from $p_\theta(\cdot\mid x)$ and returns the majority normalized answer \citep{wang2022self}. Under the same answer budget, SC spends all samples on independent full trajectories, whereas \method spends the budget on repairing or diversifying uncertain suffixes of a smaller set of draft anchors. This allows \method to reuse reliable early steps while focusing additional sampling on steps that are more likely to affect the final answer, as illustrated in Fig.~\ref{fig:overview}.

\begin{algorithm}[t]
\caption{\method: Step-Level Error-Localized Resampling}
\label{alg:method}
\small
\begin{algorithmic}[1]
\Require Problem $x$, model $p_\theta$, budget $B=n_d n_s$, localization rule $\mathrm{Loc}$
\Ensure Final answer $a^\star$
\State $\mathcal{A}\leftarrow \emptyset$ \Comment{multiset of candidate answers}
\For{$j=1,\ldots,n_d$}
    \State $\tau^{(j)}=(z^{(j)}_1,\ldots,z^{(j)}_{T_j+1})
    \sim p_\theta(\cdot\mid x)$
    \State $\mathcal{A}\leftarrow \mathcal{A}\cup\{a(\tau^{(j)})\}$ \Comment{keep the draft's own answer}
    \State $\hat{e}_j \leftarrow \mathrm{Loc}(x,\tau^{(j)})$
    \For{$m=1,\ldots,n_s-1$}
        \State $\tilde{z}^{(j,m)}
        \sim p_\theta(\cdot\mid x,z^{(j)}_{<\hat{e}_j})$
        \State $\mathcal{A}\leftarrow \mathcal{A}\cup
        \{a(z^{(j)}_{<\hat{e}_j},\tilde{z}^{(j,m)})\}$
    \EndFor
\EndFor
\State \Return
$
a^\star =
\operatorname{mode}\!\left(
\operatorname{Normalize}(\mathcal{A})
\right)
$
\end{algorithmic}
\end{algorithm}
\section{Experiments}
\label{sec:experiments}

We evaluate whether \method can outperform or achieve a comparable performance to full-trajectory self-consistency (SC) at a fixed candidate-answer budget while reducing generation cost.  The results focus on budget $B=32$ and compare rollback-resampling against Greedy@1 and SC@32.

\subsection{Experimental Setup}
\label{sec:exp_setup}

\paragraph{Benchmarks.}
We evaluate mathematical reasoning on GSM8K~\citep{cobbe2021training}, MATH500~\citep{hendrycks2021measuring}, AMC2023, AIME24, and OlympiadBench~\citep{he-etal-2024-olympiadbench}, covering tasks from grade-school arithmetic to competition-level mathematics. 
We further evaluate multi-hop question answering on HotpotQA~\citep{yang-etal-2018-hotpotqa} and 2WikiMultiHopQA~\citep{ho-etal-2020-constructing}, which require composing evidence across multiple supporting facts. 
For efficiency, we randomly sample 500 examples from the HotpotQA validation set of 7,405 examples and 1,000 examples from the 2WikiMultiHopQA validation set of 12.6K examples.

\paragraph{Models.}
The model set contains \texttt{Qwen-2.5-3B-Instruct}, \texttt{Llama-3.2-3B-Instruct}, and \texttt{Qwen-2.5-7B-Instruct}. These models span various architectures and scales, allowing us to test whether error-localized suffix resampling is robust beyond a single backbone.

\paragraph{Baselines and variants.}
The primary baseline is self-consistency (SC)~\citep{wang2022self}, which samples \(B=32\) complete chain-of-thought trajectories independently and selects the most frequent normalized final answer. We also report greedy chain-of-thought decoding (CoT), which generates one deterministic reasoning trajectory and serves as a low-cost reference point. In Section~\ref{sec:adaptive_results}, we additionally compare against adaptive reasoning methods, including Adaptive Consistency (AC)~\citep{aggarwal2023lets}, Early-Stopping Self-Consistency (ESC)~\citep{li2024escape}, and Difficulty-Adaptive Self-Consistency (DSC)~\citep{wang2025penny}. Details for their hyperparameters can be found in Appendix~\ref{app:hyperparameter}. For the main comparison, \method uses the same candidate-answer budget as SC@32. We consider two lightweight variants of the localization signal: \textsc{NLL-signal} and \textsc{PRM-signal}, both described in Section~\ref{sec:localization_signals}. 

Unless otherwise specified, we use thresholds \(0.1\) for \textsc{NLL-signal} and \(0.05\) for \textsc{PRM-signal}, without dataset-specific tuning.
Lower thresholds trigger a rollback on modest drops of the step-level signal, so more drafts are rewritten since an earlier step; this favors accuracy over token saving.
These thresholds are shared across all datasets and models; a held-out calibration on training questions (Appendix~\ref{app:threshold_calibration}) shows that every threshold in a wide grid matches or exceeds the average accuracy of SC@32 while saving 20--44\% of tokens.

\paragraph{Evaluation metrics.}
For all datasets, we report answer-level accuracy after answer extraction and normalization. For mathematical reasoning, extracted answers are compared against the reference answer using exact or symbolic equivalence when applicable. For multi-hop QA, we use normalized exact match. In Table~\ref{tab:main_results}, we report SC@32 accuracy once for each model-dataset pair and report \method by its accuracy change over SC,
\begin{equation}
    \Delta\mathrm{Acc}
    =
    \mathrm{Acc}(\method)-\mathrm{Acc}(\mathrm{SC}@32),
\end{equation}
where positive values indicate that \method outperforms SC. We measure generation efficiency by the token saving relative to SC@32,
\begin{equation}
    \eta
    =
    100\left(
    1-\frac{\mathrm{Tok/Q}(\method)}
            {\mathrm{Tok/Q}(\mathrm{SC}@32)}
    \right),
\end{equation}
where \(\mathrm{Tok/Q}\) denotes the average number of generated tokens per question. In Table~\ref{tab:adaptive_complement_results}, we further report path expansion (PExp) \citep{jindal2026path}, the average number of sampled reasoning continuations per question, and
\begin{equation}
    \mathrm{G/1kT}
    =
    1000 \cdot
    \frac{
    \mathrm{Acc}(m)-\mathrm{Acc}(\mathrm{CoT})
    }{
    \mathrm{Tok/Q}(m)
    },
\end{equation}
which measures the accuracy gain over greedy CoT per 1k generated tokens. Higher \(\eta\) and G/1kT indicate better inference efficiency.

\paragraph{Implementation details.}
\label{sec:implementation_details}
Unless otherwise specified, we use temperature $0.7$ and top-$p$ $0.95$ for sampling.
All experiments are run on compute nodes with two AMD EPYC 9005-series CPUs, and NVIDIA RTX PRO 6000 Blackwell Server Edition GPUs (96\,GB each); each run uses 4 GPUs with CUDA 12.8, PyTorch 2.7.0, and vLLM 0.9.2.

\subsection{Main Results}
\label{sec:main_results}
Table~\ref{tab:main_results} presents the main comparison between \method and SC@32 under the same candidate-answer budget \(B=32\). 
The main table reports the default \method configuration with \(n_d=16\) draft trajectories and \(n_s=2\) candidate answers per draft (the draft's own answer and one localized suffix sample); additional budget allocations are reported in Section~\ref{sec:ablation} and Appendix~\ref{app:additional_results}. 
For each model--dataset pair, we report the SC@32 accuracy once, followed by the accuracy change \(\Delta\mathrm{Acc}\), token saving \(\eta\), and G/1kT of \method with \textsc{PRM-signal} and \textsc{NLL-signal} localization; the \textit{Avg.} row gives the mean over the seven datasets weighted by their number of questions. 

\begin{table*}[t]
\centering
\small
\setlength{\tabcolsep}{4pt}
\caption{
Main comparison between self-consistency and \method on \texttt{Llama-3B}, \texttt{Qwen-3B}, and \texttt{Qwen-7B} under the same candidate-answer budget \(B=32\).
\method uses \((n_d,n_s)=(16,2)\) with thresholds \(\delta_{\mathrm{NLL}}=0.1\) and \(\delta_{\mathrm{PRM}}=0.05\) for all three models.
We report SC@32 accuracy once per dataset, followed by the accuracy change \(\Delta\mathrm{Acc}\) (percentage points), token saving \(\eta\) (\%) and localization overhead \(\rho_{16}\) (\%) of \method.
\(\rho_{16}\) is the share of wall-clock GPU time spent scoring the 16 drafts, relative to drafting, scoring, and suffix generation combined.
G/1kT is the accuracy gain over greedy CoT per 1k generated tokens; the highest G/1kT in each row is in bold.
Positive, zero, and negative accuracy changes are shown in green, gray, and dark red, respectively.
\textit{Avg.} is the mean over the seven datasets weighted by their number of questions (4,166 in total).
Five-seed results are reported in Appendix~\ref{app:random_seed}.
}

\label{tab:main_results}
\begin{tabular}{l r | rr | rrrr | rrrr}
\toprule
 & \multicolumn{1}{c|}{Greedy} & \multicolumn{2}{c|}{SC@32} & \multicolumn{4}{c|}{\textsc{NLL-signal}} & \multicolumn{4}{c}{\textsc{PRM-signal}} \\
\cmidrule(lr){2-2}\cmidrule(lr){3-4}\cmidrule(lr){5-8}\cmidrule(lr){9-12}
Dataset & Acc & Acc & G/1kT & $\Delta$Acc & $\eta$ & G/1kT & $\rho_{16}$ & $\Delta$Acc & $\eta$ & G/1kT & $\rho_{16}$ \\
\midrule

\multicolumn{12}{l}{\texttt{Llama-3B}} \\
\midrule
GSM8K & 77.41 & 88.02 & 1.488 & \textcolor{gray}{+0.00} & 34.47 & 2.270 & 9.14 & \textcolor{ForestGreen}{+0.99} & 34.77 & \textbf{2.494} & 10.47 \\
MATH500 & 42.20 & 53.40 & 0.691 & \textcolor{ForestGreen}{+0.60} & 24.22 & 0.961 & 8.23 & \textcolor{ForestGreen}{+1.20} & 20.81 & \textbf{0.966} & 9.18 \\
AMC2023 & 15.66 & 33.73 & 0.705 & \textcolor{red!65!black}{-3.61} & 14.92 & 0.663 & 7.27 & \textcolor{ForestGreen}{+2.41} & 10.83 & \textbf{0.896} & 8.46 \\
AIME24 & 7.78 & 15.56 & \textbf{0.273} & \textcolor{red!65!black}{-2.22} & 12.41 & 0.222 & 7.08 & \textcolor{red!65!black}{-2.22} & 9.72 & 0.216 & 9.29 \\
OlympiadBench & 14.84 & 23.00 & 0.299 & \textcolor{ForestGreen}{+0.45} & 13.81 & 0.366 & 7.42 & \textcolor{ForestGreen}{+1.48} & 10.68 & \textbf{0.395} & 8.36 \\
HotpotQA & 35.00 & 50.20 & 2.606 & \textcolor{red!65!black}{-1.60} & 31.80 & \textbf{3.419} & 7.37 & \textcolor{red!65!black}{-2.00} & 7.94 & 2.458 & 3.41 \\
2Wiki & 32.80 & 44.60 & 1.903 & \textcolor{gray}{+0.00} & 30.41 & \textbf{2.734} & 8.82 & \textcolor{red!65!black}{-0.40} & 8.55 & 2.009 & 3.91 \\
\cmidrule(lr){1-12}
\textit{Avg.} & 44.53 & 55.74 & 1.392 & \textcolor{red!65!black}{-0.17} & 27.74 & \textbf{1.978} & 8.38 & \textcolor{ForestGreen}{+0.36} & 18.66 & 1.769 & 7.48 \\
\midrule
\multicolumn{12}{l}{\texttt{Qwen-3B}} \\
\midrule
GSM8K & 85.14 & 91.21 & 0.638 & \textcolor{red!65!black}{-0.45} & 34.64 & 0.903 & 11.00 & \textcolor{ForestGreen}{+0.38} & 38.37 & \textbf{1.100} & 14.32 \\
MATH500 & 61.80 & 72.80 & 0.601 & \textcolor{red!65!black}{-0.40} & 25.46 & 0.777 & 10.52 & \textcolor{ForestGreen}{+0.40} & 25.60 & \textbf{0.837} & 11.39 \\
AMC2023 & 33.73 & 43.37 & 0.370 & \textcolor{ForestGreen}{+2.41} & 24.37 & \textbf{0.612} & 9.79 & \textcolor{ForestGreen}{+2.41} & 15.07 & 0.545 & 10.06 \\
AIME24 & 4.44 & 6.67 & 0.077 & \textcolor{gray}{+0.00} & 18.47 & 0.094 & 9.77 & \textcolor{ForestGreen}{+1.11} & 8.66 & \textbf{0.126} & 10.90 \\
OlympiadBench & 26.26 & 34.87 & 0.328 & \textcolor{red!65!black}{-0.45} & 20.74 & \textbf{0.393} & 10.42 & \textcolor{gray}{+0.00} & 14.83 & 0.386 & 10.49 \\
HotpotQA & 23.20 & 33.00 & 1.282 & \textcolor{ForestGreen}{+2.60} & 22.41 & \textbf{2.091} & 7.56 & \textcolor{ForestGreen}{+1.20} & 7.66 & 1.559 & 5.00 \\
2Wiki & 26.90 & 43.70 & 2.231 & \textcolor{ForestGreen}{+0.10} & 25.82 & \textbf{3.026} & 7.41 & \textcolor{red!65!black}{-0.90} & 8.52 & 2.308 & 5.11 \\
\cmidrule(lr){1-12}
\textit{Avg.} & 48.63 & 58.71 & 1.026 & \textcolor{ForestGreen}{+0.12} & 27.15 & \textbf{1.434} & 9.52 & \textcolor{ForestGreen}{+0.17} & 21.07 & 1.266 & 9.86 \\
\midrule
\multicolumn{12}{l}{\texttt{Qwen-7B}} \\
\midrule
GSM8K & 91.13 & 93.63 & 0.304 & \textcolor{gray}{+0.00} & 38.55 & 0.494 & 13.71 & \textcolor{ForestGreen}{+0.45} & 41.26 & \textbf{0.611} & 10.38 \\
MATH500 & 71.00 & 77.20 & 0.360 & \textcolor{gray}{+0.00} & 30.03 & 0.514 & 11.26 & \textcolor{ForestGreen}{+2.00} & 28.53 & \textbf{0.666} & 7.79 \\
AMC2023 & 46.99 & 50.60 & 0.140 & \textcolor{ForestGreen}{+2.41} & 27.44 & 0.322 & 10.40 & \textcolor{ForestGreen}{+6.02} & 17.98 & \textbf{0.455} & 8.95 \\
AIME24 & 6.67 & 12.22 & \textbf{0.186} & \textcolor{red!65!black}{-2.22} & 21.66 & 0.143 & 10.78 & \textcolor{red!65!black}{-2.22} & 10.39 & 0.125 & 6.95 \\
OlympiadBench & 37.54 & 44.66 & 0.287 & \textcolor{ForestGreen}{+0.30} & 25.27 & \textbf{0.399} & 11.49 & \textcolor{red!65!black}{-0.30} & 18.04 & 0.335 & 7.58 \\
HotpotQA & 37.40 & 47.40 & 1.496 & \textcolor{ForestGreen}{+2.40} & 32.29 & \textbf{2.740} & 7.55 & \textcolor{ForestGreen}{+2.20} & 7.79 & 1.980 & 2.96 \\
2Wiki & 44.20 & 54.80 & 1.590 & \textcolor{ForestGreen}{+0.90} & 35.53 & \textbf{2.675} & 8.47 & \textcolor{ForestGreen}{+0.60} & 7.82 & 1.822 & 3.53 \\
\cmidrule(lr){1-12}
\textit{Avg.} & 59.63 & 66.25 & 0.754 & \textcolor{ForestGreen}{+0.55} & 33.32 & \textbf{1.263} & 10.93 & \textcolor{ForestGreen}{+0.82} & 22.80 & 1.014 & 6.98 \\
\bottomrule
\end{tabular}%
\end{table*}

\paragraph{\method matches SC@32 accuracy on average with fewer generated tokens.}
The \textit{Avg.} rows of Table~\ref{tab:main_results} summarize the comparison over all 4,166 questions. With \textsc{PRM-signal}, \method improves average accuracy over SC@32 by \(0.36\), \(0.17\), and \(0.82\) percentage points on Llama-3B, Qwen-3B and Qwen-7B, respectively, while generating \(21.07\%\), \(18.66\%\), and \(22.80\%\) fewer tokens. With \textsc{NLL-signal}, it saves even more tokens, stays under \(0.2\) points difference with SC@32 on the two 3B models (\(+0.12\) and \(-0.17\)), and is \(0.55\) points above SC@32 on Qwen-7B. Both signals also raise the accuracy gained per generated token: on Llama-3B, G/1kT increases from 1.392 for SC@32 to 1.769 with \textsc{PRM-signal} and 1.978 with \textsc{NLL-signal}.

Per dataset, \textsc{PRM-signal} reduces generated tokens in all 21 model--dataset pairs (7.66--41.26\%), improves accuracy in 14 of them, ties in 1, and is lower in 6. On MATH500, it improves Qwen-3B, Llama-3B, and Qwen-7B by \(+0.40\), \(+1.20\), and \(+2.00\) points while saving \(25.60\%\), \(20.81\%\), and \(28.53\%\) tokens, respectively. Notably, although the PRM is trained primarily for mathematical process supervision, it also provides useful step-level reliability estimates for logical and multi-hop reasoning: on HotpotQA, \textsc{PRM-signal} improves Qwen-3B and Qwen-7B by \(+1.20\) and \(+2.20\) points but lowers Llama-3B by \(2.00\) points, which suggests that \textsc{PRM-signal} can capture some reasoning-quality drops beyond purely arithmetic errors, but less reliably than on mathematical reasoning.

Meanwhile, \textsc{NLL-signal} provides a complementary low-cost localization signal requiring no external model: it achieves larger token savings than \textsc{PRM-signal} on 17 of the 21 pairs and improves accuracy of Qwen-3B and Qwen-7B on two multi-hop QA datasets. We factor this in that \textsc{NLL-signal} tends to localize errors less aggressively, often rolling back closer to the final answer and therefore resampling shorter suffixes. This yields stronger token savings, but may leave more earlier reasoning errors unedited compared to \textsc{PRM-signal}.

\paragraph{Localization overhead is small relative to generation.}
The additional cost of computing localization signals remains modest compared with the cost of generating complete reasoning trajectories. The relative localization overhead \(\rho_{16}\) in Table~\ref{tab:main_results}, i.e., the share of the wall-clock time of \method spent scoring the drafts, averages \(8.38\%\), \(9.52\%\), and \(10.93\%\) for \textsc{NLL-signal} and \(7.48\%\), \(9.86\%\), and \(6.98\%\) for \textsc{PRM-signal} on Llama-3B, Qwen-3B, and Qwen-7B, respectively. Thus, the token savings achieved by \method are not offset by a large scoring cost. 

\paragraph{\method benefits from reusable early reasoning steps.}
The gains of accuracy suggest that many incorrect draft trajectories are not entirely unhelpful since they often contain correct early steps. Instead of discarding these steps as SC does, \method keeps the reliable part of each draft and resamples only from the predicted error region. 
This design is especially effective on datasets where errors often arise in later reasoning stages, such as GSM8K, MATH500, and AMC2023, where \textsc{PRM-signal} improves accuracy for all three models while saving 10.83--41.26\% of generated tokens.
The consistent token reduction across all settings further indicates that localized resampling avoids a large amount of redundant early steps generation.

\paragraph{Limitations across harder datasets.}
On harder competition-style benchmarks such as AIME24 and OlympiadBench, the gains are less stable, especially for Llama-3B. This suggests that some failures cannot be corrected by resampling only a localized suffix, since they may require revising earlier reasoning decisions or exploring a substantially different solution path. 

Overall, these results support the central premise of \method: under a fixed answer budget, additional test-time compute can be spent more efficiently by resampling from a localized likely error step instead of regenerating every candidate from scratch.

\subsection{\method as a Complement to Adaptive Reasoning}
\label{sec:adaptive_results}

\begin{table*}[t]
\centering
\scriptsize
\setlength{\tabcolsep}{2.4pt}
\caption{
Combining \method with adaptive reasoning methods using \ourmodel and maximum budget \(B=32\).
CoT denotes greedy chain-of-thought decoding. For \method, the candidate pool is built from \(n_d=16\) draft anchors and \(n_s=2\) candidate answers per draft (the draft's own answer and one localized suffix sample), using \textsc{PRM-signal}. PExp~\citep{jindal2026path} is the average number of consumed candidate expansions per question. For SC/AC/ESC/DSC, each expansion is a full trajectory. For \method, each expansion is a draft trajectory or a localized suffix continuation. Tok. is the number of generated tokens per question, and G/1kT is the accuracy gain over CoT per 1k generated tokens.
Within each pair of a baseline and its \method counterpart, bold numbers indicate the better value, with higher being better for Acc. and G/1kT and lower being better for Tok. and PExp.
}
\label{tab:adaptive_complement_results}
\resizebox{0.9\textwidth}{!}{%
\begin{tabular}{l | rrrr | rrrr | rrrr}
\toprule
\multirow{2}{*}{Method} & \multicolumn{4}{c|}{\mathfive} & \multicolumn{4}{c|}{\gsm} & \multicolumn{4}{c}{HotpotQA} \\
\cmidrule(lr){2-5}\cmidrule(lr){6-9}\cmidrule(lr){10-13}
 & Acc. & PExp & Tok. & G/1kT & Acc. & PExp & Tok. & G/1kT & Acc. & PExp & Tok. & G/1kT \\
\midrule
CoT & 61.80 & 1.00 & 577 & 0.000 & 85.14 & 1.00 & 295 & 0.000 & 23.20 & 1.00 & 240 & 0.000 \\
SC & 72.80 & \textbf{32.00} & 18{,}335 & 0.600 & 91.21 & \textbf{32.00} & 9{,}506 & 0.638 & 33.00 & \textbf{32.00} & 7{,}643 & 1.282 \\
\method (ours) & \textbf{73.20} & \textbf{32.00} & \textbf{13{,}620} & \textbf{0.837} & \textbf{91.59} & \textbf{32.00} & \textbf{5{,}860} & \textbf{1.100} & \textbf{34.20} & \textbf{32.00} & \textbf{7{,}056} & \textbf{1.559} \\
\addlinespace[0.25em]
AC & \textbf{72.80} & 16.57 & 11{,}386 & 0.966 & 91.13 & 10.84 & 3{,}509 & 1.707 & 33.00 & \textbf{28.09} & 6{,}792 & 1.443 \\
\method+ AC & 72.60 & \textbf{15.46} & \textbf{8{,}749} & \textbf{1.234} & \textbf{92.34} & \textbf{9.81} & \textbf{2{,}166} & \textbf{3.326} & \textbf{34.80} & 28.11 & \textbf{6{,}286} & \textbf{1.845} \\
\addlinespace[0.25em]
ESC & \textbf{73.00} & 17.14 & 11{,}951 & 0.937 & 91.21 & 10.47 & 3{,}446 & 1.760 & 32.80 & \textbf{29.29} & 7{,}061 & 1.360 \\
\method+ ESC & 72.20 & \textbf{15.41} & \textbf{9{,}075} & \textbf{1.146} & \textbf{92.27} & \textbf{8.97} & \textbf{2{,}138} & \textbf{3.334} & \textbf{35.00} & 29.41 & \textbf{6{,}567} & \textbf{1.797} \\
\addlinespace[0.25em]
DSC & \textbf{71.80} & 13.33 & 9{,}022 & 1.108 & 90.90 & 8.66 & 2{,}790 & 2.065 & 32.60 & \textbf{21.48} & 5{,}161 & 1.821 \\
\method+ DSC & 71.60 & \textbf{12.40} & \textbf{7{,}068} & \textbf{1.387} & \textbf{92.12} & \textbf{7.90} & \textbf{1{,}815} & \textbf{3.843} & \textbf{34.00} & 21.71 & \textbf{4{,}851} & \textbf{2.226} \\
\bottomrule
\end{tabular}%
}
\end{table*}

Table~\ref{tab:adaptive_complement_results} shows that \method can be directly combined with adaptive reasoning methods. AC, ESC, and DSC originally operate on a pool of independently generated full trajectories and decide how many candidates to consume. We instead replace this full-trajectory pool with the \method suffix pool: \method first generates draft anchors and localized suffix continuations, and the adaptive policy is then applied to these suffix candidates. Therefore, \method reduces the token cost of each candidate, while AC/ESC/DSC further reduces the number of consumed candidates.

These two reductions compound. Relative to each adaptive method alone, adding \method cuts generated tokens by 22--24\% on \mathfive, 35--38\% on \gsm, and 6--7\% on HotpotQA. Accuracy also improves in all six comparisons on \gsm and HotpotQA, by 1.1--2.2 points, and G/1kT improves in all nine; on \mathfive, accuracy is only 0.2--0.8 points lower. The most efficient combination, \method+DSC on \gsm, uses 5.2\(\times\) fewer tokens than SC@32 while being 0.9 points more accurate.

Overall, the combination improves the accuracy-cost trade-off over both standard SC and adaptive reasoning alone, showing that \method is a complementary intra-trajectory allocation strategy rather than a replacement for adaptive inference.

\section{Analysis and Discussions}
\label{sec:ablation}

\paragraph{Different Process Reward Model Size.}

Table~\ref{tab:prm_size_math500} in Appendix~\ref{app:additional_results} studies whether \method relies on a large PRM for error localization. We fix the generator to Qwen2.5-3B and evaluate on \mathfive, comparing \texttt{Skywork-o1-Open-PRM-Qwen-2.5-1.5B}~\citep{he_2024_16998085} with \texttt{Qwen2.5-Math-PRM-7B} under the same \(B=32\) candidate-answer budget.

The results show that \method does not require a PRM larger than the generator. With the 1.5B Skywork PRM, \method still outperforms SC@32 in the \(8\times4\) setting, improving accuracy by \(+0.36\) points while saving \(23.8\%\) generation tokens, although the accuracy is \(0.36\) points below SC@32 at \(16\times2\). The 7B Qwen PRM gives stronger localization and achieves a higher accuracy gain of \(+0.72\) points at \(16\times2\), but 1.5B PRM already preserves a substantial portion of the benefit with lower scoring overhead (13.7ms vs. 19.7ms per draft sample).

This suggests that the key requirement is not PRM scale, but whether the localization signal is informative. Even lightweight signals, such as \textsc{NLL-signal} or a smaller PRM, can support effective suffix resampling when they identify useful rollback points.

\paragraph{Drafts versus suffix samples.}

Table~\ref{tab:appendix_qwen_prm} in Appendix~\ref{app:additional_results} studies how to allocate the same candidate budget \(B=32\) between draft anchors and suffix samples. Suffix-heavy settings such as \(1\times32\) and \(2\times16\) save the most tokens and $2\times16$ attains the highest G/1kT on three out of the seven datasets, since many candidates share the same short early steps. However, they are less stable in accuracy because a small number of flawed draft prefixes can limit recovery.

Draft-heavy settings such as \(8\times4\) and \(16\times2\) are more robust because they provide more diverse reasoning prefixes. $8\times4$ improves over SC@32 on all five math benchmarks, and $16\times2$ improves on GSM8K, MATH500, AMC2023, AIME24, and HotpotQA and matches it on OlympiadBench, while still saving tokens. These exhibit that \(n_d\) and \(n_s\) control an accuracy-efficiency trade-off: suffix-heavy settings are more cost-oriented, whereas draft-heavy settings are more accuracy-oriented.

\paragraph{Alternative rollback rules.}
Any rule that maps a draft to a rollback point can serve as \(\mathrm{Loc}\) in Algorithm~\ref{alg:method}, including signal-free rules such as random, fixed-position, and final-step rollback, and PRM-based rules from prior backtracking methods \citep{singh2025improving,tan2025adaptive}. Appendix~\ref{app:rollback_rules} compares them with drafting, suffix resampling, and majority voting held fixed. Prefix reuse and suffix resampling yield most of the token saving under any rule, and late rules such as final-step resampling are strong efficiency baselines; on HotpotQA, final-step resampling is both more accurate and cheaper than \textsc{PRM-signal}. Localization helps most when suffix samples dominate the vote: at \(4\times8\), \textsc{PRM-signal} is 2.06 points more accurate than final-step resampling over all seven datasets.

\section{Conclusion}
We presented \method, a step-level error-localized resampling method for efficient test-time reasoning. Instead of regenerating every candidate from scratch, \method keeps the reliable prefix of each draft and resamples only from a likely first error located by lightweight \textsc{NLL-signal} or \textsc{PRM-signal}. Across seven mathematical reasoning and multi-hop QA benchmarks and three models, \method matches the average accuracy of SC@32 under the same answer budget while generating up to 41.26\% fewer tokens, and it further reduces the cost of adaptive methods such as AC, ESC, and DSC. 
These results suggest that \emph{where} test-time compute is spent within a trajectory is a useful axis of efficiency, complementary to deciding \emph{how many} trajectories to sample. 

\section{Limitations and Broader Impact}\label{app:limitation}
\paragraph{Limitations. }
\method has limitations that leave room for future work. First, we
focus on two representative localization signals, \textsc{NLL-signal} and \textsc{PRM-signal}. Although they already show clear benefits, exploring a broader set of even lighter-weight signals may further improve the efficiency of \method. Second, our experiments mainly focus on mathematical reasoning and multi-hop question answering. While \method performs well in these settings, its generalizability to other domains such as code generation or long-context summarization remains to be further studied.

\paragraph{Broader impact.  }\label{app:broader_impacts}
More efficient reasoning can make stronger inferences accessible at a lower cost, which is a positive outcome for scientific and educational use. At the same time, cheaper reasoning can also make capable systems easier to deploy at scale. We therefore believe that research on efficient reasoning should ideally improve both performance and transparency about where the gains come from.

\bibliographystyle{plainnat}
\bibliography{refs}

\newpage
\appendix
\section*{Appendix}

\section{Step Decomposition Examples}
\label{app:step_examples}

Our method assumes that model outputs can be decomposed into reasoning steps. In practice, the tested instruction-tuned models usually produce line-delimited or numbered reasoning.

For Qwen2.5-3B-Instruct, steps are often separated by blank lines, i.e., ``\texttt{\textbackslash n\textbackslash n}'', and frequently begin with explicit numbers such as ``1.'', ``2.'', etc. We use these boundaries to parse trajectories into steps.

\subsection{GSM8K Example}

The following is a representative Qwen2.5-3B-Instruct output on GSM8K.

\begin{quote}
\small
Sure, let's break down the problem step by step.

1. \textbf{Determine the total number of eggs laid per day:} \\
Janet's ducks lay 16 eggs per day.

2. \textbf{Calculate the number of eggs eaten for breakfast:} \\
Janet eats 3 eggs for breakfast each morning.

3. \textbf{Calculate the number of eggs used for baking muffins:} \\
Janet uses 4 eggs to bake muffins for her friends.

4. \textbf{Calculate the total number of eggs used:} \\
\(3 + 4 = 7\).

5. \textbf{Calculate the number of eggs remaining:} \\
\(16 - 7 = 9\).

6. \textbf{Determine the revenue:} \\
\(9 \times 2 = 18\).

So, Janet makes \(\boxed{18}\) dollars every day at the farmers' market.
\end{quote}

This output is naturally segmented into six reasoning steps plus the final answer. The blank-line and numbered-step structure makes the parsing deterministic.

\subsection{Multi-hop QA Example}

The same behavior appears in general QA tasks. For example, on a HotpotQA-style multihop question, Qwen2.5-3B-Instruct produces:

\begin{quote}
\small
Let's break down the question and answer it step by step:

1. \textbf{Identify the key information}: The question is about Curt Menefee, Michael Anthony Strahan, and their NFL show.

2. \textbf{Determine the Fox network's NFL show}: The Fox network has a show featuring Curt Menefee and Michael Anthony Strahan, called ``Fox NFL Sunday.''

3. \textbf{Identify the co-host}: Michael Anthony Strahan is one of the co-hosts of this show.

4. \textbf{Find the team with which Strahan spent his entire 15-year career}: Michael Anthony Strahan played for the New York Giants.

5. \textbf{Conclusion}: The team is the New York Giants.

The answer is: The New York Giants.
\end{quote}

We apply the same segmentation rule across datasets and models. The final answer sentence is treated as an additional special step.

\section{Oracle First-Error Annotation Protocol}
\label{app:first_error_annotation}

For diagnostic analysis, we construct reference first-error annotations using GPT-5.1 as an external verifier. Given an incorrect model-generated trajectory, we prompt GPT-5.1 to identify the earliest step that contains a clear arithmetic, formula, or reasoning error. These annotations are used only for oracle and diagnostic experiments, not for training or tuning \method.

The following prompt is used for GSM8K-style math problems.

{\small
\begin{verbatim}
You are a rigorous math reasoning verifier.

Question:
{question}

Correct final answer: {gold_answer}

The model produced the following step-by-step solution
(which arrives at a WRONG final answer):

{numbered_steps}

Judgment criteria (follow strictly):
- Focus ONLY on mathematical operations. For each step, check whether the
  arithmetic (addition, subtraction, multiplication, percentages, etc.) is
  performed correctly on the numbers it uses.
- A step is CORRECT if: (a) its arithmetic is right, AND (b) the numbers it
  uses come from the problem statement or from a preceding step's result.
  It does not matter what the step calls or labels the result.
- A step is INCORRECT only if: (a) it computes an arithmetic operation wrong
  (e.g. 2+3=6), OR (b) it applies a formula/percentage to a number that
  neither the problem statement nor any preceding step provided as the
  appropriate base for that operation.
- Steps that merely restate the problem, outline a plan, set up notation,
  or compute simple sums of previously stated numbers are correct even if
  the label/name they give the result is imprecise.
- Do NOT mark a step wrong for imprecise wording or naming.
- Do NOT mark a step wrong just because it propagates an earlier mistake.

Task:
1. Walk through every step. For each, decide correct/incorrect by the
   criteria above.
2. If you find an incorrect step, report the FIRST one and explain in one
   sentence what error it introduces.
3. If NO step contains a clear arithmetic or formula error by the criteria
   above, set first_error_step to -1.

Output STRICT JSON:
{
  "first_error_step": <int, 1-indexed; or -1 if no clear error found>,
  "total_steps": {len(steps)},
  "reason": "<one sentence explaining the error, or 'no clear error found'>"
}
\end{verbatim}
}

We applied this verifier to all 16 sampled drafts per question for Qwen2.5-3B-Instruct on GSM8K and MATH500. To assess annotation quality, we manually inspected 100 randomly sampled annotations and found no obvious misjudgments under the stated criteria. We therefore use these annotations as oracle first-error labels for diagnostic experiments. We emphasize that these labels are used only for analysis and oracle studies; the proposed test-time method does not require GPT-5.1 annotations.

\begin{table*}[t]
\centering
\small
\caption{
Example of a first-error annotation. The model predicts \(180\), while the correct answer is \(540\). The annotated first error is Step 3 under our parsed-step indexing, corresponding to the model's numbered Step 2.
}
\label{tab:first_error_example}
\begin{tabular}{p{0.18\linewidth}p{0.76\linewidth}}
\toprule
Field & Content \\
\midrule

Question
& James decides to run 3 sprints 3 times a week. He runs 60 meters each sprint. How many total meters does he run a week? \\

\midrule
Gold answer
& \(540\) \\

\midrule
Predicted answer
& \(180\) \\

\midrule
Model response
&
\begin{minipage}[t]{0.82\linewidth}
To determine the total number of meters James runs in a week, we can break down the problem into smaller, manageable steps.

\vspace{0.5em}
\textbf{1. Calculate the total meters run per sprint:}

James runs 60 meters per sprint.

\vspace{0.5em}
\textbf{2. Calculate the total meters run per week for one sprint:}

James runs 3 sprints each week. Therefore, the total meters run in one week for one sprint is:
\(60 \, \text{meters/sprint} \times 3 \, \text{sprints/week} = 180 \, \text{meters/week}\).

\vspace{0.5em}
\textbf{3. Calculate the total meters run per week for all sprints:}

Since we have already calculated that James runs 180 meters per week for one sprint, we can state the final total meters run per week as:
\(180 \, \text{meters/week}\).

\vspace{0.5em}
Putting the final answer in the required format:
\(\boxed{180}\)
\end{minipage}
\\

\midrule
Annotated first error
& Parsed Step 3, corresponding to the model's numbered Step 2. \\

\midrule
Reason
& The model incorrectly interprets ``3 sprints 3 times a week'' as only 3 sprints per week instead of \(3 \times 3 = 9\) sprints per week. Therefore, it uses the wrong base for the weekly-distance calculation. \\

\bottomrule
\end{tabular}
\end{table*}

\section{First-Error Annotation Example}
\label{app:first_error_example}

Table~\ref{tab:first_error_example} shows an annotated example.

\section{First-Error Localization Diagnostics}
\label{app:localization_details}

In this section, we evaluate whether lightweight signals can approximate the oracle first-error positions used in Section~\ref{sec:oracle_error_localization}. For each incorrect trajectory, we have an annotated first-error position \(e\). Given a localization signal, we predict a first-error position \(\hat e\) and compare it against \(e\). We also include a baseline, \textsc{Random}, that samples a rollback position randomly from the available reasoning steps in the trajectory. This baseline tests whether the signal-based methods do better than selecting a resampling point without using any information from the trajectory.

For each trajectory with annotated first-error position \(e\), we evaluate the predicted position \(\hat e\) using the following four metrics.

\paragraph{I. Exact.}
\textsc{Exact} measures whether the predicted position exactly matches the annotated first-error step:
\begin{equation}
    \hat e = e.
\end{equation}
\paragraph{II. Prev-1.}
\textsc{Prev-1} measures whether the predicted position is either the annotated first-error step or the immediately preceding step:
\begin{equation}
    \hat e \in \{\max(1,e-1), e\}.
\end{equation}
This metric captures whether the predicted resampling point is close enough to cover the first error while avoiding substantial wasted computation on already-correct early reasoning steps. When \(\hat e=e\), resampling starts exactly at the first error; when \(\hat e=e-1\), only one additional valid step is resampled.

\paragraph{III. Not-Later.}
\textsc{Not-Later} measures whether the predicted position occurs before or exactly at the first error:
\begin{equation}
    \hat e \le e.
\end{equation}
This metric is important because resampling from a position before or at the true first error includes the erroneous step in the resampled suffix. However, it should be interpreted together with pre-error distance: a very early prediction can satisfy \textsc{Not-Later} while wasting computation by resampling many already-correct early reasoning steps. Hence, we also introduce the following metric.

\paragraph{IV. Mean Pre-Error Distance.}
\textsc{Mean Pre-Error Distance} measures how far the predicted position is before the annotated first-error step, counting only predictions that occur before or at the true first error:
\begin{equation}
    e - \hat e
    \quad \text{for trajectories with } \hat e \le e.
\end{equation}
A lower value indicates that the predicted resampling point is closer to the first error while still covering it. This metric distinguishes useful early localization from overly conservative (i.e., too late) predictions that restart too much of the reasoning trajectory.

\begin{table}[t]
\centering
\small
\caption{
Full first-error localization diagnostics. We compare lightweight predicted error positions \(\hat e\) against annotated first-error positions \(e\). Prev-1 measures whether the prediction is exactly the first-error step or one step before it, indicating accurate localization without excessive early resampling.
}
\label{tab:first_error_localization_full}
\begin{tabular}{lcccc}
\toprule
Method & Exact & Prev-1 & Not-Later & Mean Pre-Error Dist. \\
\midrule
\textsc{PRM-signal} \((\delta=0.05)\) & 22.3\% & 32.8\% & 52.3\% & 1.07 \\
\textsc{PRM-signal} \((\delta=0.10)\) & 21.4\% & 31.2\% & 48.6\% & 0.98 \\
\textsc{PRM-signal} \((\delta=0.20)\) & 20.5\% & 29.5\% & 45.4\% & 0.93 \\
\textsc{NLL-signal} \((\delta=0.10)\) & 15.5\% & 26.4\% & 43.5\% & 1.23 \\
\textsc{NLL-signal} \((\delta=0.20)\) & 11.5\% & 19.4\% & 29.2\% & 0.83 \\
\textsc{NLL-signal} \((\delta=0.40)\) & 8.7\% & 13.8\% & 19.5\% & 0.20 \\
Random & 11.9\% & 25.0\% & 56.4\% & 1.95 \\
\bottomrule
\end{tabular}
\end{table}

\paragraph{Discussion.}
Table~\ref{tab:first_error_localization_full} shows that both lightweight signals contain information about where reasoning first becomes unreliable. \textsc{PRM-signal} with a threshold of $\delta=$0.05 achieves the best Exact and Prev-1 scores. For the same threshold, \(52.3\%\) Not-Later rate indicates that the predicted point often falls no later than the annotated first-error step, making it possible to rewrite the erroneous step itself.  Meanwhile, its Mean Pre-Error Distance remains at only \(1.07\), explaining that it typically wastes about one correct step on average. This shows that an external process-level scorer can identify local reasoning degradation more accurately and place the resampling point close to the oracle first error without excessive early resampling. \textsc{NLL-signal} is weaker than \textsc{PRM-signal}, but still outperforms the Random baseline in all metrics except for Not-Later, exhibiting that sudden increases in token-level uncertainty may also provide a meaningful, though less precise, signal of reasoning failure.

\section{Full Oracle Error-Localized Resampling Results}
\label{app:oracle_error_localization_full}

Table~\ref{tab:oracle_error_localization_full} reports the full oracle error-localized resampling sweep. In this oracle setting, resampling starts from the annotated first-error position. When no clear first error is annotated, we fall back to resampling from the final-answer step. Here, \(n_d\) denotes the number of draft trajectories, and \(n_s\) denotes the number of candidate answers obtained from each draft.

These results provide additional evidence for the motivating observation in Section~\ref{sec:oracle_error_localization}: if the first error can be accurately localized, suffix resampling can reuse useful early reasoning steps, reduce token usage, and achieve comparable or even better accuracy than full self-consistency.

\begin{table}[h]
\centering
\small
\caption{
Full oracle error-localized resampling results. Resampling starts from annotated first-error positions. Savings are computed relative to SC@32 on the same dataset.
}
\label{tab:oracle_error_localization_full}
\begin{tabular}{llccc}
\toprule
Dataset & Method & Accuracy & Tokens / Q & Savings \\
\midrule
GSM8K & SC@32 & 91.21\% & 9506 & -- \\
& Oracle error-localized resampling \(n_d=1,n_s=32\) & 88.48\% & 1580 & 83.4\% \\
& Oracle error-localized resampling \(n_d=2,n_s=16\) & 89.46\% & 1863 & 80.4\% \\
& Oracle error-localized resampling \(n_d=4,n_s=8\) & 91.43\% & 2381 & 74.9\% \\
& Oracle error-localized resampling \(n_d=8,n_s=4\) & 91.89\% & 3427 & 64.0\% \\
& Oracle error-localized resampling \(n_d=16,n_s=2\) & 92.27\% & 5449 & 42.7\% \\
\midrule
MATH500 & SC@32 & 72.80\% & 18335 & -- \\
& Oracle error-localized resampling \(n_d=1,n_s=32\) & 68.00\% & 6852 & 62.6\% \\
& Oracle error-localized resampling \(n_d=2,n_s=16\) & 68.60\% & 7524 & 59.0\% \\
& Oracle error-localized resampling \(n_d=4,n_s=8\) & 71.40\% & 8441 & 54.0\% \\
& Oracle error-localized resampling \(n_d=8,n_s=4\) & 72.40\% & 9815 & 46.5\% \\
& Oracle error-localized resampling \(n_d=16,n_s=2\) & 73.00\% & 12615 & 31.2\% \\
\bottomrule
\end{tabular}
\end{table}

\section{Threshold Calibration on Held-out Data}
\label{app:threshold_calibration}

The main experiments use a single threshold per signal for all datasets and models, \(\delta_{\mathrm{NLL}}=0.1\) and \(\delta_{\mathrm{PRM}}=0.05\). To examine how sensitive \method is to this choice, we calibrate both thresholds on held-out training questions.

\paragraph{Protocol.}
We sample 100 questions from the MATH training set~\citep{hendrycks2021measuring} and 100 from the HotpotQA training set~\citep{yang-etal-2018-hotpotqa}. We run \method with \ourmodel in the \(16\times2\) setting under two random seeds and sweep each threshold over the grid in Table~\ref{tab:threshold_calibration}. Each threshold is scored by the macro average of MATH and HotpotQA accuracy, and the highest-scoring threshold is selected, with ties broken in favor of fewer generated tokens.

\paragraph{Results.}
Table~\ref{tab:threshold_calibration} shows that \method is robust to the threshold: every value in the grid matches or exceeds the macro score of SC@32 (62.00) while saving 19.67--44.37\% of generated tokens. Larger thresholds flag fewer steps, so more drafts keep their full reasoning chain and resample only the final answer step (Section~\ref{sec:localization_signals}); token savings therefore increase monotonically with \(\delta\). As \(\delta\) increases, HotpotQA accuracy tends to increase, while MATH accuracy stays within 1.5 points of SC@32.

The calibration selects \(\delta_{\mathrm{NLL}}=0.5\) and \(\delta_{\mathrm{PRM}}=0.8\), both larger than our defaults. On this held-out set, the defaults match MATH accuracy at the SC@32 level (80.00) and save 25.25\% and 21.43\% of tokens, whereas the selected thresholds save 42.58\% and 38.92\% with a higher macro score. Our defaults thus sit at the low-saving end of the grid rather than at the calibrated optimum, which suggests that the gains in Section~\ref{sec:main_results} do not rely on a favorable choice of \(\delta\).

\begin{table}[t]
\centering
\small
\caption{
Threshold calibration on held-out training questions. Results are averaged over two seeds on 100 MATH and 100 HotpotQA training questions, using \ourmodel in the \(16\times2\) setting. Macro is the mean of MATH accuracy and HotpotQA, and \(\eta\) is the token saving relative to SC@32. The selected threshold for each signal is shown in bold, and \(^\dagger\) marks the default used in the main experiments.
}
\label{tab:threshold_calibration}
\begin{tabular}{lcccc}
\toprule
\(\delta\) & MATH & HotpotQA & Macro & \(\eta\) (\%) \\
\midrule
SC@32 & 80.00 & 44.00 & 62.00 & -- \\
\midrule
\multicolumn{5}{l}{\textsc{NLL-signal}} \\
0.05 & 80.00 & 45.00 & 62.50 & 19.67 \\
0.1\(^\dagger\) & 80.00 & 45.50 & 62.75 & 25.25 \\
0.15 & 79.00 & 45.50 & 62.25 & 29.90 \\
0.2 & 79.00 & 46.00 & 62.50 & 33.86 \\
0.3 & 78.50 & 48.50 & 63.50 & 38.19 \\
0.4 & 79.00 & 49.50 & 64.25 & 41.12 \\
\textbf{0.5} & \textbf{78.50} & \textbf{50.00} & \textbf{64.25} & \textbf{42.58} \\
0.6 & 78.50 & 49.50 & 64.00 & 43.38 \\
0.7 & 78.50 & 49.00 & 63.75 & 43.83 \\
0.8 & 78.50 & 49.50 & 64.00 & 44.09 \\
0.9 & 78.50 & 49.50 & 64.00 & 44.23 \\
\midrule
\multicolumn{5}{l}{\textsc{PRM-signal}} \\
0.025 & 80.00 & 48.50 & 64.25 & 19.73 \\
0.05\(^\dagger\) & 80.00 & 47.50 & 63.75 & 21.43 \\
0.075 & 80.50 & 48.00 & 64.25 & 22.49 \\
0.1 & 80.50 & 47.00 & 63.75 & 23.15 \\
0.15 & 79.50 & 48.00 & 63.75 & 24.41 \\
0.2 & 79.50 & 48.50 & 64.00 & 25.62 \\
0.3 & 79.50 & 50.00 & 64.75 & 27.75 \\
0.4 & 79.50 & 48.50 & 64.00 & 29.94 \\
0.5 & 78.50 & 49.50 & 64.00 & 31.98 \\
0.6 & 79.50 & 49.00 & 64.25 & 34.26 \\
0.7 & 79.50 & 49.50 & 64.50 & 36.59 \\
0.75 & 79.50 & 49.50 & 64.50 & 37.71 \\
\textbf{0.8} & \textbf{80.00} & \textbf{50.00} & \textbf{65.00} & \textbf{38.92} \\
0.9 & 80.00 & 49.50 & 64.75 & 41.05 \\
1.0 & 79.00 & 49.50 & 64.25 & 44.37 \\
\bottomrule
\end{tabular}
\end{table}

\section{Comparison with Alternative Rollback Rules}
\label{app:rollback_rules}

The framework of \method does not fix how the rollback point is chosen. For a given allocation \((n_d,n_s)\), drafting, suffix resampling, and majority voting are consistent across rules, and only the localization rule \(\mathrm{Loc}\) of Section~\ref{sec:method} changes. This lets us separate the effect of localization from the effect of reusing prefixes and resampling suffixes. All results in this section use \texttt{Qwen-3B}, the sampling settings of Section~\ref{sec:exp_setup}, and \(B=32\) answers per question.

\paragraph{Signal-free rules.}
These rules ignore the content of the draft. \textsc{Random} rolls back to a step drawn uniformly at random from the draft's reasoning steps, as in Table~\ref{tab:first_error_localization_full}. \textsc{Fixed}-\(q\) rolls back to the step at a fixed fraction \(q\in\{25\%,50\%,75\%\}\) of the draft's reasoning steps. \textsc{Final-step} sets \(\hat e_j=T_j+1\) for every draft: it keeps the whole reasoning chain and resamples only the final-answer step. This is also the fallback of \textsc{NLL-signal} and \textsc{PRM-signal} when no step crosses the threshold, so both signals approach \textsc{Final-step} as \(\delta\) grows.

\paragraph{Rules inspired by prior backtracking methods.}
Several test-time methods revise a reasoning trace from a step flagged as erroneous. We keep only their rule for choosing that step, compute it with the same PRM as \textsc{PRM-signal} (\texttt{Qwen2.5-Math-PRM-7B}), and leave the rest of \method unchanged.
\begin{itemize}
    \item \textsc{Min-advantage} is inspired by \citet{singh2025improving}. The original method identifies backtracking points based on changes in a process verifier's predictions and performs multiple rounds of revision using a verifier conditioned on previous attempts. Our version rolls back to the step with the most negative advantage \(r_t-r_{t-1}\), i.e., \(\hat e_j=\arg\max_t \Delta^{\mathrm{PRM}}_t\). Unlike \textsc{PRM-signal}, it rolls back in every draft, however small the largest drop is.
    \item \textsc{Absolute-PRM} is inspired by AR-Sampling~\citep{tan2025adaptive}. AR-Sampling scores each step with a PRM during generation. When the score falls below a fixed threshold, it keeps the step and appends a trigger sentence asking the model to rethink from that step. Our version rolls back to the first step whose PRM score is below \(0.8\), i.e., \(\hat e_j=\min\{t:r_t<0.8\}\), and otherwise falls back to \(T_j+1\). It replaces the relative score drop of \textsc{PRM-signal} with an absolute score threshold.
\end{itemize}

\begin{table}[t]
\centering
\small
\caption{
Rollback rules within \method over all seven datasets (\texttt{Qwen-3B}, \((n_d,n_s)=(16,2)\), \(B=32\), single run). Acc is the accuracy over all 4,166 questions and \(\Delta\)Acc is its change from SC@32 in points. \(\eta\) is the token saving relative to SC@32, averaged over the seven datasets with equal weight; it therefore differs from the question-weighted average in Table~\ref{tab:main_results}. All PRM-based rules use \texttt{Qwen2.5-Math-PRM-7B}. The highest accuracy and savings are in bold.
}
\label{tab:rollback_rules_all}
\begin{tabular}{l rrr}
\toprule
Rollback rule & Acc & $\Delta$Acc & $\eta$ \\
\midrule
SC@32 (no rollback) & 58.71 & -- & -- \\
\midrule
\multicolumn{4}{l}{\textit{Signal-free}} \\
\textsc{Random} & 58.47 & \textcolor{red!65!black}{-0.24} & 20.40 \\
\textsc{Fixed}-25\% & 58.62 & \textcolor{red!65!black}{-0.09} & 5.29 \\
\textsc{Fixed}-50\% & 58.59 & \textcolor{red!65!black}{-0.12} & 16.41 \\
\textsc{Fixed}-75\% & 58.76 & \textcolor{ForestGreen}{+0.05} & 33.66 \\
\textsc{Final-step} & 58.76 & \textcolor{ForestGreen}{+0.05} & \textbf{44.07} \\
\midrule
\multicolumn{4}{l}{\textit{Inspired by prior backtracking methods}} \\
\textsc{Min-advantage}~\citep{singh2025improving} & 58.47 & \textcolor{red!65!black}{-0.24} & 15.81 \\
\textsc{Absolute-PRM}~\citep{tan2025adaptive} & 58.76 & \textcolor{ForestGreen}{+0.05} & 17.89 \\
\midrule
\multicolumn{4}{l}{\textit{Localization signals (ours)}} \\
\textsc{NLL-signal} (\(\delta_{\mathrm{NLL}}=0.1\)) & 58.83 & \textcolor{ForestGreen}{+0.12} & 24.56 \\
\textsc{PRM-signal} (\(\delta_{\mathrm{PRM}}=0.05\)) & \textbf{58.88} & \textcolor{ForestGreen}{+0.17} & 16.96 \\
\bottomrule
\end{tabular}
\end{table}

\paragraph{Results over all seven datasets.}
Table~\ref{tab:rollback_rules_all} compares all rules at \((n_d,n_s)=(16,2)\) over the 4,166 questions of Table~\ref{tab:main_results}. First, every rule saves tokens relative to SC@32 (5.29--44.07\%) while staying within 0.25 points of its accuracy. This suggests that much of the token saving in \method comes from reusing prefixes and resampling suffixes, regardless of the rollback rule. Second, rules that roll back late are strong efficiency baselines: \textsc{Fixed}-75\% and \textsc{Final-step} are 0.05 points above SC@32 while saving 33.66\% and 44.07\% of tokens. Compared with \textsc{Final-step}, \textsc{NLL-signal} and \textsc{PRM-signal} improve accuracy by only 0.07 and 0.12 percentage points, respectively. Their token-saving rates are 19.51 and 27.11 percentage points lower, respectively. Third, the two signals reach the highest accuracy among all rules. \textsc{NLL-signal} is above \textsc{Random}, \textsc{Fixed}-25\%, \textsc{Fixed}-50\%, \textsc{Min-advantage}, and \textsc{Absolute-PRM} on both accuracy and saving. Among the PRM-based rules, \textsc{PRM-signal} is above \textsc{Min-advantage} on both axes (\(+0.41\) points of accuracy and \(+1.15\) points of saving), and it is 0.12 points more accurate than \textsc{Absolute-PRM} while saving 0.93 points fewer tokens. Unlike \textsc{Min-advantage}, \textsc{PRM-signal} rolls back only when the score drop reaches \(\delta\), so drafts without a clear failure keep their whole reasoning chain.

\begin{table}[t]
\centering
\small
\setlength{\tabcolsep}{4pt}
\caption{
Five-seed comparison of rollback rules on \mathfive and HotpotQA (\texttt{Qwen-3B}, \(B=32\)). Acc is the mean over five seeds \(\pm\) the sample standard deviation across seeds, and \(\eta\) is the token saving relative to SC@32. \textsc{PRM-signal} uses \(\delta_{\mathrm{PRM}}=0.05\) with either \texttt{Skywork-o1-Open-PRM-Qwen-2.5-1.5B} (1.5B) or \texttt{Qwen2.5-Math-PRM-7B} (7B). For reference, SC@32 reaches \(72.84\pm0.17\) on \mathfive over the same seeds and 33.00 on HotpotQA (Table~\ref{tab:main_results}). The highest mean accuracy in each column is in bold.
}
\label{tab:rollback_rules_seeds}
\begin{tabular}{l rr rr rr}
\toprule
 & \multicolumn{2}{c}{4\(\times\)8} & \multicolumn{2}{c}{8\(\times\)4} & \multicolumn{2}{c}{16\(\times\)2} \\
\cmidrule(lr){2-3}\cmidrule(lr){4-5}\cmidrule(lr){6-7}
Rollback rule & Acc & $\eta$ & Acc & $\eta$ & Acc & $\eta$ \\
\midrule
\multicolumn{7}{l}{\mathfive} \\
\midrule
\textsc{Random} & \(70.16\pm0.82\) & 34.8 & \(71.88\pm0.87\) & 29.9 & \(72.60\pm0.51\) & 19.9 \\
\textsc{Fixed}-25\% & \(71.48\pm0.61\) & 11.0 & \(72.32\pm0.30\) & 9.6 & \(72.60\pm0.25\) & 6.4 \\
\textsc{Fixed}-50\% & \(71.08\pm0.36\) & 28.9 & \(71.60\pm0.55\) & 24.7 & \(72.16\pm0.72\) & 16.6 \\
\textsc{Fixed}-75\% & \(69.16\pm1.32\) & 56.1 & \(71.52\pm0.67\) & 48.1 & \(72.20\pm0.45\) & 32.2 \\
\textsc{Final-step} & \(68.08\pm1.54\) & 73.9 & \(71.24\pm0.50\) & 63.4 & \(72.04\pm0.77\) & 42.4 \\
\textsc{NLL-signal} (\(\delta=0.2\)) & \(69.08\pm0.91\) & 53.2 & \(71.08\pm0.91\) & 40.6 & \(71.96\pm0.79\) & 33.0 \\
\textsc{PRM-signal} (1.5B) & \(71.88\pm0.50\) & 36.0 & \(\mathbf{73.20}\pm0.74\) & 23.8 & \(72.48\pm0.59\) & 24.6 \\
\textsc{PRM-signal} (7B) & \(\mathbf{72.40}\pm0.28\) & 39.1 & \(\mathbf{73.20}\pm0.37\) & 26.8 & \(\mathbf{73.56}\pm0.79\) & 25.9 \\
\midrule
\multicolumn{7}{l}{HotpotQA} \\
\midrule
\textsc{Random} & \(32.92\pm1.32\) & 37.1 & \(33.48\pm0.61\) & 32.4 & \(33.60\pm0.69\) & 21.7 \\
\textsc{Fixed}-25\% & \(32.16\pm0.74\) & 6.3 & \(33.48\pm0.46\) & 5.5 & \(33.48\pm0.11\) & 3.8 \\
\textsc{Fixed}-50\% & \(31.88\pm0.98\) & 29.6 & \(32.28\pm0.59\) & 25.3 & \(33.04\pm0.78\) & 17.0 \\
\textsc{Fixed}-75\% & \(32.00\pm0.86\) & 65.7 & \(33.32\pm1.14\) & 56.4 & \(34.12\pm1.19\) & 37.7 \\
\textsc{Final-step} & \(\mathbf{34.80}\pm0.98\) & 81.2 & \(\mathbf{35.72}\pm0.70\) & 69.7 & \(\mathbf{35.20}\pm0.61\) & 46.5 \\
\textsc{NLL-signal} (\(\delta=0.2\)) & \(33.48\pm0.64\) & 56.2 & \(33.68\pm0.86\) & 48.4 & \(33.96\pm0.77\) & 32.3 \\
\textsc{PRM-signal} (1.5B) & \(32.88\pm0.84\) & 45.0 & \(33.84\pm0.88\) & 38.7 & \(34.40\pm1.06\) & 25.9 \\
\textsc{PRM-signal} (7B) & \(32.16\pm0.56\) & 17.7 & \(32.88\pm0.36\) & 15.0 & \(33.72\pm1.00\) & 10.1 \\
\bottomrule
\end{tabular}
\end{table}

\paragraph{Five-seed results on \mathfive and HotpotQA.}
Table~\ref{tab:rollback_rules_seeds} repeats the comparison over five seeds for all three allocations. On \mathfive, \textsc{PRM-signal} with the 7B PRM has the highest mean accuracy at every allocation (tied with the 1.5B PRM at 8\(\times\)4). It exceeds the best signal-free rule by 0.88--0.96 percentage points. At each allocation, this gap exceeds the seed standard deviation of either method. The most accurate signal-free rule, \textsc{Fixed}-25\%, saves only 6.4--11.0\% of tokens. With the 1.5B PRM, \textsc{PRM-signal} is also above every signal-free rule at 4\(\times\)8 and 8\(\times\)4. \textsc{NLL-signal} is less accurate than most signal-free rules on \mathfive, consistent with its weaker localization in Table~\ref{tab:first_error_localization_full}.

On HotpotQA, \textsc{Final-step} is the most accurate rule at every allocation and also saves the most tokens, so neither signal improves on it for this task. Among the remaining rules, \textsc{NLL-signal} is the most accurate at 4\(\times\)8, while \textsc{PRM-signal} with the 1.5B PRM is the most accurate at 8\(\times\)4 and 16\(\times\)2. With the math-specific 7B PRM, however, \textsc{PRM-signal} is less accurate than \textsc{Random} at 4\(\times\)8 and 8\(\times\)4. The strong performance of \textsc{Final-step} cannot be explained by its 16 drafts alone. SC@16 is 1.4 and 1.2 percentage points below SC@32 on \mathfive and HotpotQA, respectively. In contrast, \textsc{Final-step} at 16\(\times\)2 is only 0.80 percentage points below SC@32 on \mathfive and 2.20 percentage points above it on HotpotQA.

\begin{table}[t]
\centering
\small
\setlength{\tabcolsep}{5pt}
\caption{
\textsc{Final-step} versus \method at each allocation over all seven datasets (\texttt{Qwen-3B}, single run). Acc and \(\eta\) are defined as in Table~\ref{tab:rollback_rules_all}; SC@32 reaches 58.71\%. The last two rows use larger thresholds, which send more drafts to the final-step fallback. The highest accuracy in each column is in bold.
}
\label{tab:rollback_allocation}
\begin{tabular}{l rr rr rr}
\toprule
 & \multicolumn{2}{c}{4\(\times\)8} & \multicolumn{2}{c}{8\(\times\)4} & \multicolumn{2}{c}{16\(\times\)2} \\
\cmidrule(lr){2-3}\cmidrule(lr){4-5}\cmidrule(lr){6-7}
Rollback rule & Acc & $\eta$ & Acc & $\eta$ & Acc & $\eta$ \\
\midrule
\textsc{Final-step} & 56.53 & 76.87 & 58.23 & 66.10 & 58.76 & 44.07 \\
\midrule
\textsc{NLL-signal} (\(\delta_{\mathrm{NLL}}=0.1\)) & 57.23 & 43.20 & 58.47 & 36.73 & 58.83 & 24.56 \\
\textsc{PRM-signal} (\(\delta_{\mathrm{PRM}}=0.05\)) & \textbf{58.59} & 29.75 & 58.79 & 25.61 & 58.88 & 16.96 \\
\midrule
\textsc{NLL-signal} (\(\delta_{\mathrm{NLL}}=0.5\)) & 56.53 & 72.06 & 57.85 & 61.84 & 58.33 & 41.33 \\
\textsc{PRM-signal} (\(\delta_{\mathrm{PRM}}=0.8\)) & 58.21 & 64.53 & \textbf{59.29} & 55.03 & \textbf{58.95} & 36.76 \\
\bottomrule
\end{tabular}
\end{table}

\paragraph{When localization matters.}
Table~\ref{tab:rollback_allocation} compares \textsc{Final-step} with both signals at every allocation over all seven datasets. The accuracy advantage of localization shrinks as draft answers account for a larger share of the 32 votes: \textsc{PRM-signal} is 2.06, 0.56, and 0.12 points more accurate than \textsc{Final-step} at 4\(\times\)8, 8\(\times\)4, and 16\(\times\)2, and \textsc{NLL-signal} is 0.70, 0.24, and 0.07 points more accurate. \textsc{Final-step} saves more tokens in every case.

The final vote explains this trend. Each draft contributes its original answer and \(n_s-1\) answers from rewritten suffixes. Thus, original draft answers contribute \(n_d\) of the 32 votes, and rewritten answers contribute the remaining \(n_d(n_s-1)\). Localization changes only the rewritten answers. Let \(V_D\), \(V_R\), and \(V_{\mathrm{full}}\) denote the most frequent answer (the mode, which need not hold a majority) among the draft answers, the rewrite answers, and all 32 answers, respectively. We call a question a \emph{conflict} when \(V_D\neq V_R\). Over the 4,166 questions, \method with \textsc{PRM-signal} has 1,098, 853, and 756 conflicts at 4\(\times\)8, 8\(\times\)4, and 16\(\times\)2 (Table~\ref{tab:vote_conflicts}). The final answer \(V_{\mathrm{full}}\) matches \(V_R\) in 89.34\%, 74.33\%, and 48.02\% of these conflicts, respectively. It matches \(V_D\) in 6.92\%, 19.58\%, and 43.39\%, respectively. In the remaining conflicts, the final answer matches neither. When rewrites contribute 28 of the 32 votes, their most frequent answer wins almost all conflicts. When drafts and rewrites contribute 16 votes each, this share falls to about half. Based on these observations, we can draw that \textbf{localization determines where suffix rewriting starts, and the allocation determines how much the rewritten answers influence the final vote.}

\begin{table}[t]
\centering
\small
\setlength{\tabcolsep}{4pt}
\caption{
Draft/rewrite conflicts of \method with \textsc{PRM-signal} over the 4,166 questions (\texttt{Qwen-3B}, \(\delta_{\mathrm{PRM}}=0.05\), single run). \(V_D\), \(V_R\), and \(V_{\mathrm{full}}\) are the most frequent answers among the draft answers, the rewrite answers, and all 32 answers. A conflict is a question with \(V_D\neq V_R\); the last three columns split the conflicts by whether \(V_{\mathrm{full}}=V_R\), \(V_{\mathrm{full}}=V_D\), or neither.
}
\label{tab:vote_conflicts}
\begin{tabular}{l cc r rrr}
\toprule
Allocation & Draft votes & Rewrite votes & Conflicts & \(V_{\mathrm{full}}=V_R\) & \(V_{\mathrm{full}}=V_D\) & Neither \\
\midrule
4\(\times\)8 & 4 & 28 & 1{,}098 & 981 (89.34\%) & 76 (6.92\%) & 41 (3.73\%) \\
8\(\times\)4 & 8 & 24 & 853 & 634 (74.33\%) & 167 (19.58\%) & 52 (6.10\%) \\
16\(\times\)2 & 16 & 16 & 756 & 363 (48.02\%) & 328 (43.39\%) & 65 (8.60\%) \\
\bottomrule
\end{tabular}
\end{table}

Raising the threshold moves both signals toward \textsc{Final-step}, because more drafts fall back to resampling only the final-answer step. The last two rows of Table~\ref{tab:rollback_allocation} use \(\delta_{\mathrm{NLL}}=0.5\) and \(\delta_{\mathrm{PRM}}=0.8\). With \(\delta_{\mathrm{PRM}}=0.8\), \textsc{PRM-signal} saves 36.76--64.53\% of tokens and is still 1.68, 1.06, and 0.19 points more accurate than \textsc{Final-step}. At 8\(\times\)4 and 16\(\times\)2, it is also 0.58 and 0.24 percentage points more accurate than SC@32, respectively. It also achieves higher accuracy and greater token savings than \textsc{PRM-signal} with \(\delta_{\mathrm{PRM}}=0.05\). With \(\delta_{\mathrm{NLL}}=0.5\), the token-saving rate of \textsc{NLL-signal} is only 2.74--4.81 percentage points lower than that of \textsc{Final-step}. Its accuracy matches \textsc{Final-step} at 4\(\times\)8 and is lower at the other two allocations, consistent with its weaker localization.

\paragraph{Summary.}
Within \method, the rollback rule sets an accuracy--efficiency trade-off rather than having a single best choice. Signal-free late rollback, especially \textsc{Final-step}, gives the largest savings and is the most accurate rule on HotpotQA. Localization signals give the highest pooled accuracy, and with a 7B PRM, the highest \mathfive accuracy, and their benefit is largest when rewritten suffixes dominate the vote. The threshold \(\delta\) moves each signal along this trade-off, from early rollback toward the \textsc{Final-step} fallback.

\section{Additional Results}
\label{app:additional_results}

\begin{table*}[t]
\centering
\tiny
\setlength{\tabcolsep}{1.8pt}
\caption{Draft--suffix allocation study for \method with \textsc{PRM-signal} on \texttt{Qwen-3B} under a fixed candidate-answer budget \(B=32\). Greedy Acc and SC@32 are shown as reference points. For each \(n_d \times n_s\)  configuration, $\Delta\mathrm{Acc}$ is relative to SC@32, $\eta$ is token saving relative to SC@32, and G/1kT is accuracy gain over Greedy CoT per 1k generated tokens. The highest G/1kT in each row is bolded.}

\label{tab:appendix_qwen_prm}
\resizebox{\textwidth}{!}{%
\begin{tabular}{l r rr | rrr rrr rrr rrr rrr}
\toprule
\multirow{2}{*}{Dataset} & \multirow{2}{*}{Greedy Acc} & \multicolumn{2}{c|}{SC@32} & \multicolumn{3}{c}{1$\times$32} & \multicolumn{3}{c}{2$\times$16} & \multicolumn{3}{c}{4$\times$8} & \multicolumn{3}{c}{8$\times$4} & \multicolumn{3}{c}{16$\times$2} \\
\cmidrule(lr){3-4}\cmidrule(lr){5-7}\cmidrule(lr){8-10}\cmidrule(lr){11-13}\cmidrule(lr){14-16}\cmidrule(lr){17-19}
 & & Acc & G/1kT & $\Delta$Acc & $\eta$ & G/1kT & $\Delta$Acc & $\eta$ & G/1kT & $\Delta$Acc & $\eta$ & G/1kT & $\Delta$Acc & $\eta$ & G/1kT & $\Delta$Acc & $\eta$ & G/1kT \\
\midrule
GSM8K & 85.14 & 91.21 & 0.64 & \textcolor{red!65!black}{-0.99} & 74.60 & 2.10 & \textcolor{red!65!black}{-0.23} & 71.74 & \textbf{2.17} & \textcolor{ForestGreen}{+0.53} & 67.24 & 2.12 & \textcolor{ForestGreen}{+0.22} & 57.38 & 1.55 & \textcolor{ForestGreen}{+0.38} & 38.37 & 1.10 \\
MATH500 & 61.80 & 72.80 & 0.60 & \textcolor{red!65!black}{-2.00} & 48.74 & 0.96 & \textcolor{red!65!black}{-2.20} & 45.89 & 0.89 & \textcolor{red!65!black}{-0.40} & 42.70 & \textbf{1.01} & \textcolor{ForestGreen}{+0.60} & 36.70 & 1.00 & \textcolor{ForestGreen}{+0.40} & 25.60 & 0.84 \\
AMC2023 & 33.73 & 43.37 & 0.37 & \textcolor{gray}{+0.00} & 33.58 & 0.56 & \textcolor{ForestGreen}{+2.41} & 31.88 & \textbf{0.68} & \textcolor{ForestGreen}{+1.21} & 29.22 & 0.59 & \textcolor{ForestGreen}{+1.21} & 24.12 & 0.55 & \textcolor{ForestGreen}{+2.41} & 15.07 & 0.55 \\
AIME24 & 4.44 & 6.67 & 0.08 & \textcolor{gray}{+0.00} & 16.04 & 0.09 & \textcolor{ForestGreen}{+1.11} & 15.30 & \textbf{0.14} & \textcolor{ForestGreen}{+1.11} & 14.53 & 0.13 & \textcolor{ForestGreen}{+1.11} & 13.11 & 0.13 & \textcolor{ForestGreen}{+1.11} & 8.66 & 0.13 \\
OlympiadBench & 26.26 & 34.87 & 0.33 & \textcolor{red!65!black}{-1.64} & 28.39 & 0.37 & \textcolor{ForestGreen}{+0.59} & 26.98 & 0.48 & \textcolor{ForestGreen}{+0.14} & 26.53 & 0.45 & \textcolor{ForestGreen}{+1.33} & 22.65 & \textbf{0.49} & \textcolor{gray}{+0.00} & 14.83 & 0.39 \\
HotpotQA & 23.20 & 33.00 & 1.28 & \textcolor{red!65!black}{-0.40} & 13.93 & 1.43 & \textcolor{red!65!black}{-0.20} & 14.07 & 1.46 & \textcolor{ForestGreen}{+0.20} & 13.51 & 1.51 & \textcolor{red!65!black}{-0.20} & 11.57 & 1.42 & \textcolor{ForestGreen}{+1.20} & 7.66 & \textbf{1.56} \\
2Wiki & 26.90 & 43.70 & 2.23 & \textcolor{red!65!black}{-2.00} & 15.86 & 2.34 & \textcolor{red!65!black}{-2.00} & 16.12 & 2.34 & \textcolor{red!65!black}{-1.40} & 15.20 & \textbf{2.41} & \textcolor{red!65!black}{-1.30} & 12.94 & 2.36 & \textcolor{red!65!black}{-0.90} & 8.52 & 2.31 \\
\bottomrule
\end{tabular}%
}
\end{table*}

\begin{table*}[t]
\centering
\tiny
\setlength{\tabcolsep}{1.8pt}
\caption{Draft--suffix allocation study for \method with \textsc{NLL-signal} on \texttt{Qwen-3B} under a fixed candidate-answer budget \(B=32\). Greedy Acc and SC@32 are shown as reference points. For each \(n_d \times n_s\)  configuration, $\Delta\mathrm{Acc}$ is relative to SC@32, $\eta$ is token saving relative to SC@32, and G/1kT is accuracy gain over Greedy CoT per 1k generated tokens. The highest G/1kT in each row is bolded.}

\label{tab:appendix_qwen_nll}
\resizebox{\textwidth}{!}{%
\begin{tabular}{l r rr | rrr rrr rrr rrr rrr}
\toprule
\multirow{2}{*}{Dataset} & \multirow{2}{*}{Greedy Acc} & \multicolumn{2}{c|}{SC@32} & \multicolumn{3}{c}{1$\times$32} & \multicolumn{3}{c}{2$\times$16} & \multicolumn{3}{c}{4$\times$8} & \multicolumn{3}{c}{8$\times$4} & \multicolumn{3}{c}{16$\times$2} \\
\cmidrule(lr){3-4}\cmidrule(lr){5-7}\cmidrule(lr){8-10}\cmidrule(lr){11-13}\cmidrule(lr){14-16}\cmidrule(lr){17-19}
 & & Acc & G/1kT & $\Delta$Acc & $\eta$ & G/1kT & $\Delta$Acc & $\eta$ & G/1kT & $\Delta$Acc & $\eta$ & G/1kT & $\Delta$Acc & $\eta$ & G/1kT & $\Delta$Acc & $\eta$ & G/1kT \\
\midrule
GSM8K & 85.14 & 91.21 & 0.64 & \textcolor{red!65!black}{-4.25} & 67.97 & 0.60 & \textcolor{red!65!black}{-4.40} & 65.44 & 0.51 & \textcolor{red!65!black}{-0.69} & 60.89 & \textbf{1.45} & \textcolor{red!65!black}{-0.46} & 51.87 & 1.23 & \textcolor{red!65!black}{-0.45} & 34.64 & 0.90 \\
MATH500 & 61.80 & 72.80 & 0.60 & \textcolor{red!65!black}{-6.20} & 48.90 & 0.51 & \textcolor{red!65!black}{-8.20} & 48.02 & 0.29 & \textcolor{red!65!black}{-3.40} & 44.80 & 0.75 & \textcolor{red!65!black}{-0.60} & 37.90 & \textbf{0.91} & \textcolor{red!65!black}{-0.40} & 25.46 & 0.78 \\
AMC2023 & 33.73 & 43.37 & 0.37 & \textcolor{red!65!black}{-4.82} & 52.70 & 0.39 & \textcolor{red!65!black}{-3.61} & 49.37 & 0.46 & \textcolor{ForestGreen}{+1.21} & 42.03 & 0.72 & \textcolor{ForestGreen}{+4.82} & 35.59 & \textbf{0.86} & \textcolor{ForestGreen}{+2.41} & 24.37 & 0.61 \\
AIME24 & 4.44 & 6.67 & 0.08 & \textcolor{red!65!black}{-1.11} & 40.92 & 0.07 & \textcolor{red!65!black}{-1.11} & 34.86 & 0.06 & \textcolor{gray}{+0.00} & 32.57 & \textbf{0.11} & \textcolor{gray}{+0.00} & 26.55 & 0.10 & \textcolor{gray}{+0.00} & 18.47 & 0.09 \\
OlympiadBench & 26.26 & 34.87 & 0.33 & \textcolor{red!65!black}{-4.31} & 40.08 & 0.27 & \textcolor{red!65!black}{-3.86} & 38.06 & 0.29 & \textcolor{red!65!black}{-3.56} & 34.90 & 0.30 & \textcolor{ForestGreen}{+0.14} & 30.13 & \textbf{0.48} & \textcolor{red!65!black}{-0.45} & 20.74 & 0.39 \\
HotpotQA & 23.20 & 33.00 & 1.28 & \textcolor{ForestGreen}{+1.00} & 44.07 & \textbf{2.53} & \textcolor{gray}{+0.00} & 42.61 & 2.23 & \textcolor{ForestGreen}{+0.60} & 39.56 & 2.25 & \textcolor{ForestGreen}{+1.40} & 33.76 & 2.21 & \textcolor{ForestGreen}{+2.60} & 22.41 & 2.09 \\
2Wiki & 26.90 & 43.70 & 2.23 & \textcolor{red!65!black}{-5.10} & 49.54 & 3.08 & \textcolor{red!65!black}{-2.30} & 49.07 & \textbf{3.78} & \textcolor{red!65!black}{-1.60} & 46.24 & 3.76 & \textcolor{red!65!black}{-1.30} & 39.71 & 3.41 & \textcolor{ForestGreen}{+0.10} & 25.82 & 3.03 \\
\bottomrule
\end{tabular}%
}
\end{table*}

\begin{table*}[t]
\centering
\scriptsize
\setlength{\tabcolsep}{3.0pt}
\caption{
Effect of PRM size on \mathfive using the same Qwen2.5-3B-Instruct generator and the same candidate-answer budget \(B=32\).
We compare SC@32 with \method using \textsc{PRM-signal} localization (\(\delta_{\mathrm{PRM}}=0.05\)), where the likely error step is predicted by either \texttt{Skywork-o1-Open-PRM-Qwen-2.5-1.5B} or \texttt{Qwen2.5-Math-PRM-7B}.
Acc. is the mean over five seeds; \(\Delta\mathrm{Acc}\) and \(\eta\) are relative to SC@32. ms/sample reports \textsc{PRM-signal} wall-clock overhead per draft sample.
}
\label{tab:prm_size_math500}
\resizebox{0.7\textwidth}{!}{%
\begin{tabular}{l | l r c | rrr}
\toprule
& PRM or Baseline & ms/sample & Config & Acc. & $\Delta$Acc & $\eta$ \\
\midrule
Baseline & SC@32 & -- & -- & 72.84 & \textcolor{gray}{+0.00} & 0.0 \\
\midrule
w/ \method & \texttt{Skywork-o1-Open-PRM-Qwen-2.5-1.5B} & 13.7 & 4$\times$8 & 71.88 & \textcolor{red!65!black}{-0.96} & 36.0 \\
 & & 13.7 & 8$\times$4 & 73.20 & \textcolor{ForestGreen}{+0.36} & 23.8 \\
 & & 13.7 & 16$\times$2 & 72.48 & \textcolor{red!65!black}{-0.36} & 24.6 \\
\midrule
w/ \method & \texttt{Qwen2.5-Math-PRM-7B} & 19.7 & 4$\times$8 & 72.40 & \textcolor{red!65!black}{-0.44} & 39.1 \\
 & & 19.7 & 8$\times$4 & 73.20 & \textcolor{ForestGreen}{+0.36} & 26.8 \\
 & & 19.7 & 16$\times$2 & 73.56 & \textcolor{ForestGreen}{+0.72} & 25.9 \\
\bottomrule
\end{tabular}%
}
\end{table*}

Table~\ref{tab:appendix_qwen_prm} and \ref{tab:appendix_qwen_nll} report all available variants using the same comparison format as the main table. We include variants for \textsc{NLL-signal} and \textsc{PRM-signal} localization with $(n_d,n_s)\in\{(1,32),(2,16),(4,8),(8,4),(16,2)\}$.

\section{Licenses for Existing Assets}
\label{app:licenses}

Table~\ref{tab:asset_licenses} summarizes the existing assets used in this work. 
We use these assets only for research evaluation and do not redistribute modified versions of the datasets or model checkpoints.

\begin{table}[t]
\centering
\small
\caption{Existing assets used in this work.}
\label{tab:asset_licenses}
\begin{tabular}{p{1in} l p{1.7in} p{1.3in}}
\toprule
Asset & Type & Source / Version & License or Terms \\
\midrule
GSM8K & Dataset & OpenAI grade-school-math / HF openai/gsm8k & MIT \\
MATH / MATH500 & Dataset & Hendrycks MATH / MATH500 split & MIT \\
OlympiadBench & Dataset & OpenBMB OlympiadBench & MIT \\
HotpotQA & Dataset & HotpotQA official release & CC BY-SA 4.0 \\
2WikiMultihopQA & Dataset & Official 2WikiMultihopQA release & Apache 2.0 \\
Qwen2.5 models & Model checkpoint & HuggingFace Qwen release & Apache 2.0 or Qwen license, depending on model size \\
Llama 3.2 models & Model checkpoint & Meta Llama 3.2 release & Llama 3.2 Community License \\
\bottomrule
\end{tabular}
\end{table}

\section{Hyperparameter Settings}
\label{app:hyperparameter}

\paragraph{Early-Stopping Methods.}
We combine StepBack with three early-stopping criteria from the self-consistency literature:
\begin{itemize}
    \item \textbf{AC} (Adaptive Consistency): Stops sampling when the Beta posterior confidence that the current plurality answer is correct exceeds $\theta_{\text{ac}} = 0.99$.

    \item \textbf{ESC} (Early Stopping Consistency): Stops when all answers within a sliding window of size $w = 5$ are unanimous.

    \item \textbf{DSC} (Difficulty-Adaptive Self-Consistency): Uses a Beta posterior with confidence threshold $\theta_{\text{dsc}} = 0.95$ for easy questions and a hard-question threshold of $0.50$; stops early when confidence is reached or abandons when confidence remains low after half the budget is consumed.
\end{itemize}

\begin{table}
\centering
\small
\setlength{\tabcolsep}{3.8pt}
\caption{
Per-seed stability on \mathfive across five random seeds using \method with \textsc{PRM-signal} (\(\delta_{\mathrm{PRM}}=0.05\)). We compare SC@32 with three configurations under the same answer budget \(B=32\).
Here, \(n_d\times n_s\) means \(n_d\) draft trajectories with \(n_s\) candidate answers per draft.
Seed columns report accuracy in percent; means are reported with the sample standard deviation across seeds.
\(\Delta_{\mathrm{SC}}\) is the paired accuracy difference from SC@32 on the same seed, in percentage points.
 \(>\) SC counts the number of seeds where \method outperforms SC@32.
}
\label{tab:seed_stability_math500}
\resizebox{0.9\columnwidth}{!}{
\begin{tabular}{l rrrrr r r r}
\toprule
Method & Seed 42 & Seed 123 & Seed 456 & Seed 789 & Seed 1024 & Mean Acc. & Mean \(\Delta_{\mathrm{SC}}\) & \(>\) SC \\
\midrule
SC@32 & 72.8 & 72.8 & 72.6 & 73.0 & 73.0 & \(72.84 \pm 0.17\) & \(+0.00 \pm 0.00\) & \texttt{--} \\
\midrule
PRM 4\(\times\)8 & 72.2 & 72.0 & 72.6 & 72.6 & 72.6 & \(72.40 \pm 0.28\) & \(-0.44 \pm 0.30\) & 0/5 \\
PRM 8\(\times\)4 & 72.6 & 73.2 & 73.2 & 73.6 & 73.4 & \(73.20 \pm 0.37\) & \(+0.36 \pm 0.33\) & 4/5 \\
PRM 16\(\times\)2 & 73.2 & 73.0 & 74.6 & 74.2 & 72.8 & \(73.56 \pm 0.79\) & \(+0.72 \pm 0.86\) & 4/5 \\
\bottomrule
\end{tabular}
}
\end{table}

\section{Random-Seed Robustness}
\label{app:random_seed}

Table~\ref{tab:seed_stability_math500} reports random-seed robustness on \mathfive across five seeds.
PRM 8\(\times\)4 and PRM 16\(\times\)2 improve over SC@32, raising the mean accuracy from \(72.84\%\) to \(73.20\%\) and \(73.56\%\), with mean paired gains of \(+0.36\) and \(+0.72\) points, while PRM 4\(\times\)8 is \(0.44\) points lower.
Both 8\(\times\)4 and 16\(\times\)2 outperform SC@32 in 4 out of 5 seeds; on the remaining seed, 8\(\times\)4 is \(0.2\) points lower and 16\(\times\)2 ties.
The gains are therefore small, but they are not caused by a single favorable seed.

\clearpage
\section*{NeurIPS Paper Checklist}

\begin{enumerate}

\item {\bf Claims}
    \item[] Question: Do the main claims made in the abstract and introduction accurately reflect the paper's contributions and scope?
    \item[] Answer: \answerYes{}{} 
    \item[] Justification: The abstract and introduction summarize the main contribution of \method as an error-localized suffix resampling method for efficient test-time reasoning. 

\item {\bf Limitations}
    \item[] Question: Does the paper discuss the limitations of the work performed by the authors?
    \item[] Answer: \answerYes{} 
    \item[] Justification: We discuss limitations in Section~\ref{app:limitation}. 

\item {\bf Theory assumptions and proofs}
    \item[] Question: For each theoretical result, does the paper provide the full set of assumptions and a complete (and correct) proof?
    \item[] Answer: \answerNA{} 
    \item[] Justification: The paper does not present formal theoretical results such as theorems or lemmas. We include only analytical discussions to provide intuition for the proposed method.

    \item {\bf Experimental result reproducibility}
    \item[] Question: Does the paper fully disclose all the information needed to reproduce the main experimental results of the paper to the extent that it affects the main claims and/or conclusions of the paper (regardless of whether the code and data are provided or not)?
    \item[] Answer: \answerYes{} 
    \item[] Justification: Section~\ref{sec:experiments} describes the evaluated datasets, models, decoding hyperparameters, sample budgets, voting procedure, and evaluation metrics needed to reproduce the main results.

\item {\bf Open access to data and code}
    \item[] Question: Does the paper provide open access to the data and code, with sufficient instructions to faithfully reproduce the main experimental results, as described in supplemental material?
    \item[] Answer: \answerYes{} 
    \item[] Justification: We use publicly available benchmarks and open-source model checkpoints. We provide a GitHub repository with the implementation and reproduction instructions.

\item {\bf Experimental setting/details}
    \item[] Question: Does the paper specify all the training and test details (e.g., data splits, hyperparameters, how they were chosen, type of optimizer) necessary to understand the results?
    \item[] Answer: \answerYes{} 
    \item[] Justification: We specify the experimental setup and implementation details in Section~\ref{sec:exp_setup}. We provide hyperparameters for methods in Section~\ref{app:hyperparameter}.

\item {\bf Experiment statistical significance}
    \item[] Question: Does the paper report error bars suitably and correctly defined or other appropriate information about the statistical significance of the experiments?
    \item[] Answer: \answerYes{} 
    \item[] Justification: In Appendix~\ref{app:random_seed}, we use different random seeds to do the experiments on the MATH 500 dataset.

\item {\bf Experiments compute resources}
    \item[] Question: For each experiment, does the paper provide sufficient information on the computer resources (type of compute workers, memory, time of execution) needed to reproduce the experiments?
    \item[] Answer: \answerYes{} 
    \item[] Justification: In Section~\ref{sec:exp_setup}, we provide implementation details.
    
\item {\bf Code of ethics}
    \item[] Question: Does the research conducted in the paper conform, in every respect, with the NeurIPS Code of Ethics \url{https://neurips.cc/public/EthicsGuidelines}?
    \item[] Answer: \answerYes{} 
    \item[] Justification: The research uses publicly available datasets and model checkpoints, does not involve private user data or human-subject experiments, and conforms to the NeurIPS Code of Ethics. We preserve anonymity in the submission and report potential societal impacts and limitations.

\item {\bf Broader impacts}
    \item[] Question: Does the paper discuss both potential positive societal impacts and negative societal impacts of the work performed?
    \item[] Answer: \answerYes{} 
    \item[] Justification: We discuss broader impacts in Section~\ref{app:broader_impacts}. \method may improve the efficiency and accessibility of reasoning-time inference, but stronger reasoning systems can also be misused or produce more persuasive incorrect outputs; we therefore emphasize careful evaluation and deployment.
    
\item {\bf Safeguards}
    \item[] Question: Does the paper describe safeguards that have been put in place for responsible release of data or models that have a high risk for misuse (e.g., pre-trained language models, image generators, or scraped datasets)?
    \item[] Answer: \answerNA{} 
    \item[] Justification:  The paper does not release a new pretrained language model, image generator, scraped dataset, or other high-risk asset.

\item {\bf Licenses for existing assets}
    \item[] Question: Are the creators or original owners of assets (e.g., code, data, models), used in the paper, properly credited and are the license and terms of use explicitly mentioned and properly respected?
    \item[] Answer: \answerYes{} 
    \item[] Justification: We provide credits and license information for the datasets and model checkpoints in Appendix~\ref{app:licenses}, and use them only for research evaluation.

\item {\bf New assets}
    \item[] Question: Are new assets introduced in the paper well documented and is the documentation provided alongside the assets?
    \item[] Answer: \answerNA{} 
    \item[] Justification: The paper does not introduce a new dataset, model checkpoint, or benchmark asset.

\item {\bf Crowdsourcing and research with human subjects}
    \item[] Question: For crowdsourcing experiments and research with human subjects, does the paper include the full text of instructions given to participants and screenshots, if applicable, as well as details about compensation (if any)? 
    \item[] Answer: \answerNA{} 
    \item[] Justification: The paper does not involve crowdsourcing experiments or human-subject studies. 

\item {\bf Institutional review board (IRB) approvals or equivalent for research with human subjects}
    \item[] Question: Does the paper describe potential risks incurred by study participants, whether such risks were disclosed to the subjects, and whether Institutional Review Board (IRB) approvals (or an equivalent approval/review based on the requirements of your country or institution) were obtained?
    \item[] Answer: \answerNA{} 
    \item[] Justification: The paper does not involve crowdsourcing or human-subject research, and therefore IRB approval is not applicable.

\item {\bf Declaration of LLM usage}
    \item[] Question: Does the paper describe the usage of LLMs if it is an important, original, or non-standard component of the core methods in this research? Note that if the LLM is used only for writing, editing, or formatting purposes and does \emph{not} impact the core methodology, scientific rigor, or originality of the research, declaration is not required.
    \item[] Answer: \answerYes{} 
    \item[] Justification: LLMs are central to the research because the proposed method operates on LLM-generated reasoning trajectories and evaluates inference-time decoding behavior. We describe the backbone models, PRM scoring models, and any LLM-assisted error annotation or analysis protocol in Section~\ref{sec:experiments} and ~\ref{app:first_error_annotation}.

\end{enumerate}

\end{document}